%% file: main.tex
\documentclass{article}
\usepackage{pulmoworld_preprint,times}
\input{math_commands.tex}

\usepackage[hypertexnames=false]{hyperref}
\usepackage{url,graphicx,amsmath,amssymb,booktabs,array,microtype,multirow,placeins,float,flafter,colortbl,tikz}
\microtypesetup{nopatch=eqnum}
\usetikzlibrary{arrows.meta,positioning}
\definecolor{tblhead}{RGB}{232,238,246}
\definecolor{tblours}{RGB}{230,244,234}
\definecolor{tblmass}{RGB}{236,242,250}
\definecolor{tblmimic}{RGB}{245,242,235}

\usepackage[capitalize,noabbrev]{cleveref}
\hypersetup{hidelinks,
 pdftitle={PulmoWorld: An Asynchronous Multimodal Medical World Model for Longitudinal Pneumonia Dynamics},
 pdfauthor={Cheng Tang; Xinyu Zhang; Jiawei Chen; Minghui Zhang; Shaocong Wei; Wei Li; Yuxing Chen; Kongcheng Zhang; Chun-Mei Feng; Nanqing Dong; Xinyuan Zhu; Ziyan Huang; Junjun He; Yirong Chen; Lingtong Huang; Min Cen; Ming Hu}}

\title{PulmoWorld: An Asynchronous Multimodal Medical World Model for Longitudinal \\ Pneumonia Dynamics}
\input{authors.tex}
\newcommand{\method}{\textsc{PulmoWorld}}

\iclrfinalcopy

\begin{document}
\raggedbottom
\maketitle
\input{sections/00_abstract}
\input{sections/01_introduction}
\input{sections/02_related_work}
\input{sections/04_method}
\input{sections/05_experiments}
\input{sections/07_limitations}

\clearpage
\bibliography{references}
\bibliographystyle{iclr2027_conference}

\clearpage
\appendix
\input{appendix/00_roadmap}
\input{appendix/a_data_privacy}
\input{appendix/b_implementation}
\input{appendix/c_additional_results}

\input{appendix/c_action_visuals}
\input{appendix/b_raw_cases}
\input{appendix/d_integrated_trajectories}
\input{appendix/d_case_evolution}
\input{appendix/d_reproducibility}
\end{document}

%% file: math_commands.tex
\usepackage{amsmath,amsfonts,bm}

\def\eqref#1{equation~\ref{#1}}

\def\1{\bm{1}}

\DeclareMathAlphabet{\mathsfit}{\encodingdefault}{\sfdefault}{m}{sl}
\SetMathAlphabet{\mathsfit}{bold}{\encodingdefault}{\sfdefault}{bx}{n}



%% file: authors.tex
\makeatletter
\gdef\@author{%
\begin{minipage}{\textwidth}\centering\normalfont
Cheng Tang$^{1,2}$, Xinyu Zhang$^{3}$, Jiawei Chen$^{1}$, Minghui Zhang$^{1,4}$,\\[2pt]
Shaocong Wei$^{1,2}$, Wei Li$^{1}$, Yuxing Chen$^{1}$, Kongcheng Zhang$^{5}$,\\[2pt]
Chun-Mei Feng$^{6}$, Nanqing Dong$^{7,1}$, Xinyuan Zhu$^{8,9}$, Ziyan Huang$^{1}$,\\[2pt]
Junjun He$^{1}$, Yirong Chen$^{1}$, Lingtong Huang$^{5}$, Min Cen$^{1,*}$, Ming Hu$^{1,*}$
\par\vspace{8pt}
{\small
$^{1}$Shanghai Artificial Intelligence Laboratory\\
$^{2}$University of the Chinese Academy of Sciences\quad $^{3}$Fudan University\\
$^{4}$Shanghai Jiao Tong University\quad $^{5}$Zhejiang University\\
$^{6}$University College Dublin\quad $^{7}$Shanghai Innovation Institute\\
$^{8}$Ant Group\quad $^{9}$University of Science and Technology of China\par}
\par\vspace{5pt}
{\small $^{*}$Co-corresponding authors:\\
\texttt{cenmin@pjlab.org.cn}, \texttt{huming@pjlab.org.cn}\par}
\end{minipage}}
\makeatother

%% file: sections/00_abstract.tex
\begin{abstract}
Pneumonia changes across respiration, inflammation, and organ function, while
the measurements that describe it arrive at different times and are often incomplete.
We introduce \method{}, a world model for longitudinal pneumonia dynamics that
connects asynchronous state estimation, recursive simulation, and
evidence-constrained planning. History-aware filtering separates state
propagation from observation-driven correction, so a measurement updates the shared belief only when it is observed.
That belief supports recursive forecasts of clinical values and modality
representations, and summaries of future prognosis.
Cap-MBRL searches imagined continuations inside an authorized action set and
replaces a frozen fallback only when a candidate meets the risk budget and a
predicted-cost margin. On the pneumonia cohort MASS, recursive forecasts are
competitive with task-trained sequence models, and simulated future states
retain prognostic information. Trained separately on public MIMIC-IV-CXR chest radiographs and clinical
trajectories, this model improves prognostic discrimination over
latest-value fusion in development and is competitive on long-horizon error on the locked
pneumonia panel. In five known-mechanism
MASS-Syn environments, Cap-MBRL lowers selection regret relative to
direct fallback execution.
\end{abstract}

%% file: sections/01_introduction.tex
\begin{figure*}[!t]
    \centering
    \includegraphics[width=\textwidth]{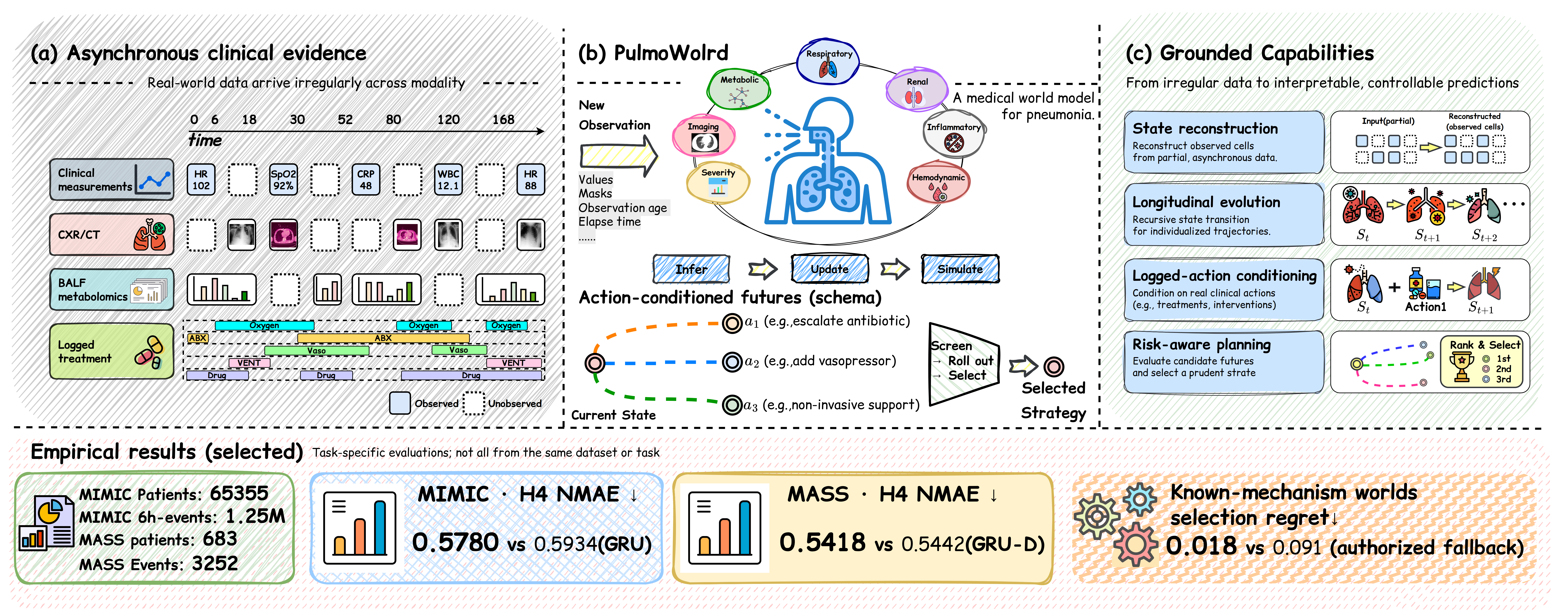}
    \caption{
    \textbf{A shared patient world for longitudinal simulation
    and evidence-constrained planning.}
    (a) Asynchronous multimodal observations inform (b) an evolving patient
    belief and (c) simulation and planning interfaces.
    The lower strip summarizes cohort forecasting and MASS-Syn selection.
    Treatment branches show imagined continuations scored in MASS-Syn.
    }
    \label{fig:teaser}
    \vspace{-3pt}
\end{figure*}

\section{Introduction}

Pneumonia evolves across respiratory function, inflammation, renal function,
and hemodynamics \citep{metlay2019cap}. A current risk score captures one aspect of this process;
longitudinal care also requires anticipating how these measurements may change
over the coming hours or days. A patient simulator must infer the
current physiological state and evolve it through time, including trajectories
conditioned on recorded or candidate care.

The evidence needed to infer this state arrives unevenly.
Clinical measurements, chest imaging, and bronchoalveolar lavage fluid (BALF)
metabolomics describe complementary aspects of a patient, but differ in
observation time and availability (\cref{fig:teaser}(a)). Combining the latest
value from each modality can mix evidence of different ages. Conversely,
requiring complete multimodal visits discards partially observed histories.
The modeling problem is to maintain a patient state between observations,
incorporate each available measurement when it arrives, and continue simulating
when no further measurements are supplied
\citep{ha2018world,lecun2022path,hafner2019learning}.

Existing medical world models address important parts of this problem.
EHRWorld learns from clinical notes and structured events, while SepsisAgent
couples a numerical clinical world model with an LLM treatment policy
\citep{mu2026ehrworld,wu2026agentifying}.
ChronoMedicalWorld models trajectories from longitudinal care records
\citep{wang2026chronomedicalworld}.
Imaging-centered models include MEWM and ORBIT
\citep{yang2025mewm,yan2026orbit}; CLARITY conditions imaging-derived dynamics
on clinical context and irregular follow-up intervals \citep{ding2026clarity}.
Our focus is the observation-to-state interface for incomplete clinical,
imaging, and molecular histories: these streams supply evidence about an
evolving patient, with available modalities changing over time.

\method{} separates state propagation from observation-driven correction.
Elapsed time advances a prior patient belief; an observed measurement corrects
that belief through a masked innovation update. A missing measurement supplies
no correction. The model tracks evidence freshness separately from state
uncertainty, and decodes future clinical values and modality representations
from recursively propagated states. This connects asynchronous evidence
assimilation to simulation through a shared state interface.

Using those simulations for planning requires an additional decision about
which action comparisons have evidential support. Recorded care conditions
the simulated trajectory.
Cap-MBRL restricts action comparisons to a fixed authorized set and evaluates
candidate continuations inside the learned world model. A candidate replaces
the frozen fallback only if it meets the risk budget and improves predicted
cost by a specified margin; otherwise, the fallback is retained.

Our contributions are threefold:
\begin{itemize}
    \setlength{\itemsep}{2pt}
    \setlength{\parskip}{0pt}

    \item \textbf{Asynchronous multimodal state estimation.}
    We develop a history-aware patient belief that integrates incomplete
    multimodal observations through state propagation and observation-driven
    correction. On MIMIC-IV-CXR development data, it improves prognostic
    discrimination relative to latest-value fusion.

    \item \textbf{Recursive patient simulation and predictive utility.}
    We model clinical and modality-specific evolution within a shared belief
    space, including treatment-conditioned trajectories. On MASS and on the
    locked MIMIC pneumonia panel, these forecasts and simulated states retain
    clinical and prognostic information.

    \item \textbf{Evidence-constrained planning.}
    We introduce Cap-MBRL, which compares authorized actions through model
    rollouts and replaces a frozen fallback only under a risk budget and a
    predicted-cost margin. In known-mechanism MASS-Syn, this selection lowers
    regret relative to executing the frozen fallback under the same authorization.
\end{itemize}

%% file: sections/02_related_work.tex
\section{Related Work}
\subsection{Medical world models}
EHR-based world models learn longitudinal patient dynamics from clinical
records. EHRWorld combines note-derived profiles with structured clinical
events \citep{mu2026ehrworld}; SepsisAgent uses a numerical physiological
world model for an LLM's propose--simulate--refine treatment workflow
\citep{wu2026agentifying}. ChronoMedicalWorld learns action-conditioned latent
transitions from clinical measurements and care communications
\citep{wang2026chronomedicalworld}.
Imaging-centered methods include MEWM and ORBIT
\citep{yang2025mewm,yan2026orbit}. CLARITY combines imaging-derived states,
clinical and genomic context, and irregular follow-up intervals for trajectory
simulation and treatment planning \citep{ding2026clarity}.
Latent imagination, learned-model search, and uncertainty-penalized offline
control supply the planning background
\citep{hafner2020dream,hafner2021dreamerv2,hafner2025dreamerv3,schrittwieser2020muzero,chua2018pets,yu2020mopo,yu2021combo}.
Sepsis-3 supplies the clinical definition, and reporting guidelines describe how
such policies should be evaluated
\citep{singer2016sepsis3,gottesman2019guidelines}.
Logged intensive-care studies then learn treatment policies from recorded
trajectories, including model-based sepsis treatment
\citep{komorowski2018aiclinician,raghu2017sepsis,raghu2018sepsis,xu2025meddreamer}.
BEHRT and Med-BERT pretrain on structured records, and LLaVA-Med aligns
biomedical images with text \citep{li2020behrt,rasmy2021medbert,li2023llavamed}.
Our task keeps incomplete clinical, imaging, and molecular evidence in one
belief, then compares authorized actions in a known-mechanism simulator.

\subsection{Multimodal patient-state modeling}
Missingness and irregular sampling are established problems in clinical
time-series modeling. GRU-D models missingness in recurrent prediction,
BRITS learns recurrent imputation, and latent ODEs describe continuous-time
latent evolution \citep{che2018recurrent,cao2018brits,rubanova2019latent}.
Continuous-time encoders and chest-radiograph datasets extend this setting
\citep{chen2018neuralode,shukla2021mtan,irvin2019chexpert}.
General-purpose and medical VLMs provide complementary image--text
representations \citep{bai2025qwen25vl,chen2024internvl,xu2025lingshu,jiang2025hulumed}.
The filter updates this belief when evidence arrives and continues the forecast
after observations stop. Experiments separate history retention, direct VLM
forecasts, and task-trained recursive dynamics.

%% file: sections/04_method.tex
\section{Method}
\label{sec:method}

\begin{figure*}[t]
\centering
\includegraphics[width=\textwidth]{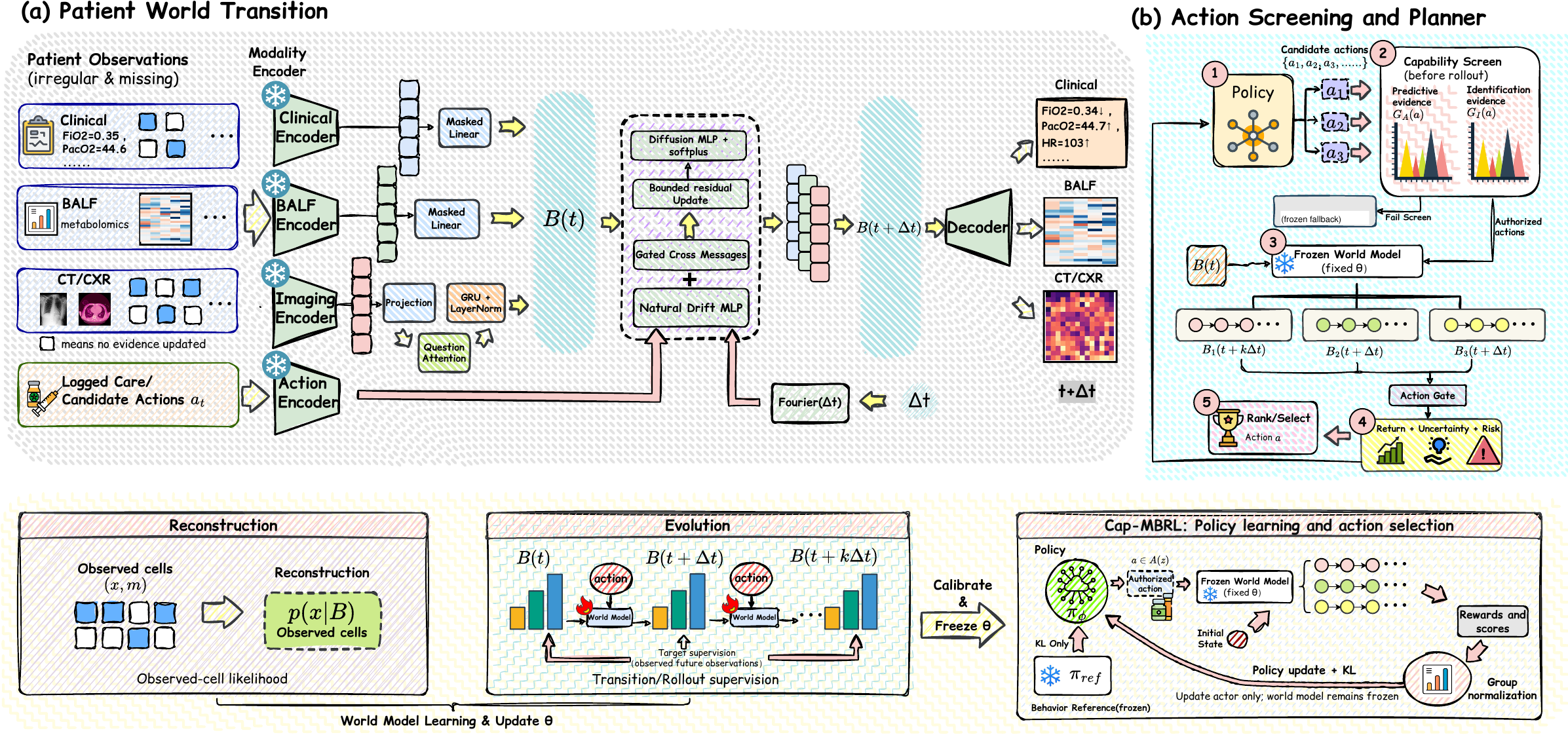}
\caption{\textbf{Pneumonia state learning and evidence-constrained planning.}
(a) Irregular clinical, imaging, and BALF observations update one patient belief,
which a shared transition then propagates.
(b) In MASS-Syn, a capability screen authorizes actions before frozen-world
rollouts; predicted return, uncertainty, and risk are used when ranking
candidates against the fallback.
Below, reconstruction and rollout losses fit the world model.}
\label{fig:method}
\end{figure*}

\method{} maintains one patient belief that integrates available observations,
evolves when measurements stop, and branches under supplied care.
The framework connects asynchronous estimation, shared-state simulation, and
evidence-constrained planning (\cref{fig:method}).

\vspace{-3pt}
\subsection{Patient state and multimodal observations}
\label{sec:problem}

We infer patient state from cohort-specific observations,
$\mathcal M_{\mathrm{MASS}}=
\{\mathrm{clinical},\mathrm{CT},\mathrm{BALF}\}$ and
$\mathcal M_{\mathrm{MIMIC}}=\{\mathrm{clinical},\mathrm{CXR}\}$.
Clinical inputs are numerical measurements; imaging and BALF metabolomics
use precomputed representations.
Let $\mathcal H_t=((o_j,m_j,\tau_j,a_{j-1}))_{j=1}^{t}$ denote observed
history, including masks, timestamps, and preceding care when available.
The belief $B_t=q_\theta(Z_t\mid\mathcal H_t)$ represents a single latent patient state $Z_t$.
Masks distinguish missing entries from measured zeros without requiring
all modalities to be observed together.
The belief-transition model $\mathcal T_\theta(B'\mid B,\Delta t,a)$
advances this state over elapsed time, optionally conditioned on care $a$.
An observation model decodes values, modality representations, and availability,
supporting evidence assimilation and simulation from the same state.

\vspace{-3pt}
\subsection{Asynchronous multimodal belief learning}
\label{sec:state-estimation}

History-aware filtering combines the state predicted from past evidence
with newly available observations. It first propagates $B_{t-1}$ to the
current timestamp, yielding $\widetilde B_t$, and then corrects this prior.
For structured filtering, the patient belief has Gaussian components
with means $\mu_t^c$ and diagonal variances.
Here $c$ indexes internal state components, not observation modalities;
dataset-specific partitions are specified in \cref{app:architecture}.
Within each modality, observation groups $v$ are encoded as
$e_t^v=E_v(m_t^v\odot o_t^v,m_t^v)$ and update their assigned component $c(v)$:
\begin{equation}
\nu_t^v=e_t^v-D_v(\widetilde\mu_t^{c(v)}),\qquad
\mu_t^c=\widetilde\mu_t^c+
\sum_{v:c(v)=c}K_t^v\odot P_v(\nu_t^v).
\label{eq:main-belief-update}
\end{equation}
Here $\nu_t^v$ is the observation innovation, and $D_v$ and $P_v$
map between state and token spaces.
The learned correction gain $K_t^v$ is zero when the group has no observed
evidence. Missing observations therefore leave the propagated state intact,
while available measurements update its mean and uncertainty.
Belief variance increases through learned diffusion during propagation and
contracts under measurement correction. Evidence reliability instead records
freshness: observed evidence resets it to one, and it decays between updates.
Thus, state uncertainty and evidence freshness serve distinct roles.
Full variance, gain, and recency equations appear in \cref{app:method-details}.

\vspace{-3pt}
\subsection{Shared-state dynamics and world-model fitting}
\label{sec:world-dynamics}

During recursive simulation, each component combines local dynamics with messages from connected components.
Supplied care branches this same transition through a state-dependent residual:
\begin{equation}
\widetilde\mu_{t+1}^c=\mu_t^c+\eta_c(\Delta t_t)
\left[F_{\mathrm{loc}}^c(\mu_t^c,T(\Delta t_t))
+\sum_{j\rightarrow c}g_{j\rightarrow c,t}M_{j\rightarrow c}(\mu_t^j)
+R_a^c(\mu_t^c,a_t,\Delta t_t)\right].
\label{eq:main-transition}
\end{equation}
$T$ encodes $\Delta t_t=\tau_{t+1}-\tau_t$; $\eta_c$ is a bounded time gain
with $\eta_c(0)=0$. Each $M_{j\rightarrow c}$ maps a source component to a
message, weighted by $g$ using state and evidence reliability.
The care residual $R_a^c$ is initialized to zero.
Removing it recovers state-only prediction.
After initialization, predicted beliefs feed subsequent transitions without
assimilating factual intermediate observations.
Decoded imaging outputs are representation forecasts rather than pixels.

World fitting grounds the source belief in observations and supervises its
recursive evolution through
\begin{equation}
\mathcal L_{\mathrm{world}}=
\lambda_{\mathrm{rec}}\mathcal L_{\mathrm{rec}}+
\lambda_{\mathrm{dyn}}\mathcal L_{\mathrm{dyn}}+
\lambda_{\mathrm{avail}}\mathcal L_{\mathrm{avail}}+
\mathcal L_{\mathrm{aux}}.
\label{eq:main-world-fitting}
\end{equation}
Reconstruction and forecasting use Gaussian negative log-likelihoods on observed
targets; availability uses binary cross-entropy. Auxiliary terms provide
alignment and consistency supervision where enabled.
Future observations supply targets, not intermediate rollout inputs.
The curriculum trains reconstruction before rollouts; the history ablation
also evaluates auxiliary reconstruction. MASS-Syn fits its transition ensemble
using one-step latent squared error.
After fitting and calibration, $\theta$ is frozen; weights and schedules
appear in \cref{app:training,app:method-details}.

\vspace{-3pt}
\subsection{Evidence-constrained planning}
\label{sec:planner}

Cap-MBRL first determines which action comparisons have evidential support,
then learns and selects within that authorized set.
Predictive evidence $G_A$ requires action-conditioned improvement with
shuffling and support checks. Identification evidence $G_I$ additionally
checks action-effect accuracy and decision regret against a known mechanism.
The authorized set $\mathcal A_{\mathrm{auth}}$ contains actions
passing both checks; the actor is normalized only over this set.
In MASS-Syn, one fixed action-channel certificate instantiates this set at every simulated state and retains the no-intervention reference. That certificate is an input to policy optimization, not an output learned by the actor.
Audit tests and thresholds are specified in \cref{app:method-details}.

Within this set, policy $\pi_\phi$ generates alternative continuations from
the same initial state through a frozen three-member world ensemble.
Its mean prediction advances the state; normalized disagreement supplies
uncertainty $u$, distinct from the Gaussian belief variance.
The score for continuation $k$ is
$S_{ik}=\sum_h\gamma^{h-1}(r_{ikh}-\kappa u_{ikh})$.
Here $i$ indexes initial states, $h$ indexes rollout steps, $\gamma$ discounts
future rewards, and $\kappa$ weights uncertainty. Reward is negative latent severity.
Scores are centered and standardized within each initial-state group to form
advantages $\widehat A_{ik}$. The actor update uses
\begin{equation}
\begin{aligned}
\mathcal L_{\mathrm{actor}}&=
-\mathbb E_{i,k}\left[\operatorname{sg}(\widehat A_{ik})
\sum_h\gamma^{h-1}\log\pi_\phi(a_{ikh}\mid B_{ikh})\right]
\\
&\quad+\lambda_{\mathrm{KL}}\widehat D_{\mathrm{KL}}(\pi_\phi,\mu_b)
+\lambda_{\rho}\mathbb E_{i,k}\left[(\bar\rho_{ik}-\bar b_{ik})_{+}\right],
\end{aligned}
\label{eq:main-policy}
\end{equation}
with a frozen behavior policy $\mu_b$. The stop-gradient applies only to the
advantage. The actor is categorical, so this term is a score-function gradient through $\log\pi_\phi$. $\bar\rho$ and $\bar b$ are discount-normalized risk and budget;
their positive gap enters the same sampled objective, and the
same budget screens the executed action.

At decision time, candidate first actions use the learned policy's
highest-probability authorized continuations; the frozen fallback follows its own policy.
For discounted severity-plus-risk cost $\widehat C$, discount-normalized risk
$\widehat\rho$, calibrated budget $\widehat b$, and fixed margin $\delta\geq0$,
eligible replacements are $\mathcal F(B)=\{a\in\mathcal A_{\mathrm{auth}}:
\widehat\rho(B,a)\leq\widehat b(B),\;
\widehat C(B,a)<\widehat C_{\mathrm{fb}}(B)-\delta\}$.
Given an authorized fallback first action $a_{\mathrm{fb}}$ with trajectory
cost $\widehat C_{\mathrm{fb}}$, execute
\begin{equation}
a_{\mathrm{exec}}(B)=
\begin{cases}
\arg\min_{a\in\mathcal F(B)}\widehat C(B,a), & \mathcal F(B)\neq\varnothing,\\
a_{\mathrm{fb}}, & \text{otherwise},
\end{cases}
\label{eq:main-selection}
\end{equation}
If no authorized fallback is available, selection abstains.
Risk combines ensemble disagreement, behavior surprisal, recursive--direct
discrepancy, and capability diagnostics. The budget constrains replacements,
not fallback risk. Only the first action is executed before replanning in MASS-Syn;
full definitions appear in \cref{app:method-details,app:planner-implementation}.

%% file: sections/05_experiments.tex
\begingroup
\setlength{\textfloatsep}{10pt plus 2pt minus 2pt}
\setlength{\floatsep}{8pt plus 2pt minus 2pt}
\setlength{\intextsep}{8pt plus 2pt minus 2pt}
\setlength{\abovecaptionskip}{6pt}
\section{Experiments}
\label{sec:experiments}
\label{sec:results}

\vspace{-3pt}
\subsection{Experimental setup}
\label{sec:experimental-setup}

\paragraph{Data and evaluation settings.}
We use two real-world cohorts and a separate known-mechanism simulator.
The two cohorts are trained and evaluated separately.
Public MIMIC-IV-CXR uses clinical measurements and chest radiographs from
65,355 patients
\citep{johnson2023mimic,johnson2019mimiccxr}.
Private MASS is a pneumonia cohort with clinical measurements, CT, and BALF
metabolomics. Locked forecasting evaluates 300 pneumonia patients drawn from
the 1,178 pneumonia cases in the 6,535-patient MIMIC test split, and 40 MASS
patients.
MIMIC uses 13 clinical variables and CXR representations. MASS uses Clinical19,
with seven compartments for the clinical groups, CT, and BALF, for development and for the
40-patient locked forecast. Both locked forecasters use the history-aware
belief. The MIMIC development comparison is against
latest-value fusion (\cref{sec:state-utility}).
MASS-Syn contains no patient records and tests planning in five
known-mechanism environments. Splits are patient-disjoint and transformations
are fitted on training data; cohort-specific protocols are in
\cref{app:data,app:architecture}.

\paragraph{Comparators.}
Forecasting controls are persistence and the training-set median.
Task-trained comparators include GRU, a causal Transformer, GRU-D,
independently adapted CLARITY, and a separately trained Qwen2.5-VL-7B SFT model.
General-purpose and medical Direct VLMs receive the cohort's numerical
input and output format. \Cref{tab:t2-main} reports the full
MIMIC comparison; the complete MASS roster is in \cref{tab:t2-complete}.
Policy comparisons separately test
unrestricted optimization, uncertainty/support penalties, authorization alone,
and risk-budgeted improvement.

\paragraph{Evaluation.}
History retention is assessed under a frozen mortality-risk probe using AUROC,
AUPRC, and Brier score, with accuracy and balanced accuracy as descriptive metrics.
Recursive forecasts advance H1--H4 without factual intermediate observations.
Macro NMAE averages errors normalized by training IQR;
$\Delta$-NMAE measures error in consecutive changes, exposing trajectories
that miss temporal variation despite plausible endpoint values.
We also report H4 direction agreement. Patient-level bootstrap $95\%$ intervals quantify uncertainty. Newly reported
paired history, MIMIC forecast, and prognosis contrasts use 5,000 resamples,
with outcome stratification for classification and fixed fitted models.
History metrics are computed per seed and then averaged.
Planning regret uses simulator-scored discounted cost relative to the best
authorized action sequence. That sequence is used only to score regret.
MASS-Syn certificates use the known mechanism as the action-effect reference
and are not a result about real MASS or MIMIC.
Metric definitions and task-specific uncertainty protocols
are in \cref{app:experiment-metrics}.

\vspace{-3pt}
\subsection{Patient-state estimation from incomplete histories}
\label{sec:state-utility}

We first compare asynchronous history filtering with latest-value fusion on 3,758 MIMIC
development patients (524 in-hospital deaths), using three seeds and one frozen
window per patient. Training both inputs with and without auxiliary masked-source
reconstruction tests whether history contributes information beyond the training objective.

\input{tables/tab_history_retention}

Asynchronous filtering improved AUROC over latest-value fusion by $0.0213$
($95\%$ paired CI $[0.0132,0.0295]$) without reconstruction
(\cref{tab:history-retention}).
With reconstruction, the AUROC, AUPRC, and Brier changes against latest-value
fusion are $+0.0381$, $+0.0778$, and $-0.0110$.
The combined comparison performed best among these four cells. The 3,758 patients are the MIMIC development set: 13 clinical variables and CXR representations, one frozen window per patient. The contrast is latest-value fusion. MASS is trained separately, so its clinical, CT, and BALF forecasts are outside the comparison reported here.

We also evaluate source completion under missing evidence (\cref{tab:t1-vlm}). This frozen history-aware MIMIC checkpoint fills 718 masked values across 271 requests (macro NMAE $0.834$); the locked forecast tables are separate three-seed means.
In development, clinical RMSE rose from $0.7688$ to $0.9504$ at 75\% source dropout
(\cref{tab:t1-main,tab:missing-evidence}).

\vspace{-3pt}
\subsection{Recursive and treatment-conditioned simulation}
\label{sec:rollout-results}

We next evaluate recursive forecasts initialized from observed evidence.
\Cref{tab:t2-main} reports the full MIMIC comparison, and
\cref{fig:forecast-comparison}a,b summarizes the MASS clinical rollouts.
On MIMIC, \method{} reached H4 NMAE $0.5780$, versus
$0.5934$ for GRU and $0.5954$ for the causal Transformer, although the
Transformer and GRU had lower H1 and H2 errors, respectively.
Relative to adapted CLARITY, H4 NMAE decreased by $0.0548$
($95\%$ paired CI $[0.0331,0.0757]$).
On MASS, mean H1--H4 NMAE was $0.557$, compared with $0.570$ for GRU-D,
with the remaining task-trained comparisons shown in \cref{fig:forecast-comparison}a.
GRU-D remained close at H4 ($0.5442$ versus $0.5418$ for \method{}).
MASS consecutive-change error was $0.498$, compared with $0.503$ for GRU-D
(\cref{fig:forecast-comparison}b).

\input{tables/tab_t2_main}

\begin{figure}[!htbp]
\centering
\includegraphics[width=\textwidth]{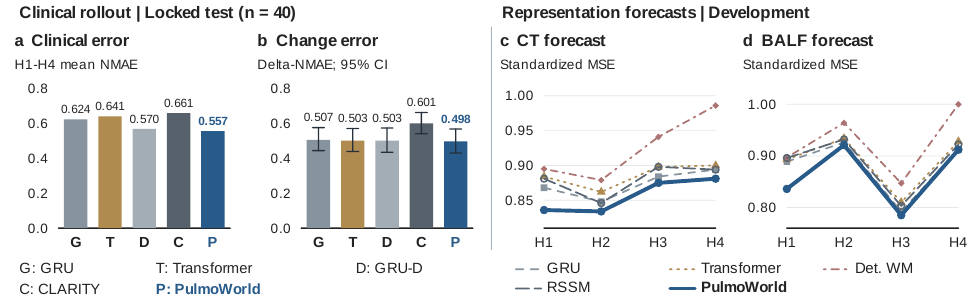}
\caption{\textbf{Recursive forecasts of clinical measurements and multimodal
representations on MASS.}
(a,b) Clinical19 locked-test mean H1--H4 NMAE and change error ($n=40$;
patient-bootstrap $95\%$ CIs in b).
(c,d) Clinical19 development CT/BALF representation MSE, not the 40-patient locked forecast
(three-seed means; non-zero axis origins).
Blue denotes \method{}; lower is better. Complete scores are in
\cref{tab:t2-complete,tab:imaging}.}
\label{fig:forecast-comparison}
\vspace{-3pt}
\end{figure}

\vspace{-2pt}
\paragraph{Multimodal evolution.}
Beyond clinical measurements, MASS development rollouts predicted
CT and BALF representations (\cref{fig:forecast-comparison}c,d).
H1 MSE was $0.836$ for both modalities, versus $0.868$ for CT
and $0.889$ for BALF with GRU
(\cref{tab:imaging}).
Together with clinical rollouts, these results show that the model simulates
multiple aspects of the patient state. Imaging outputs are representation
forecasts; the observed images in the case displays provide visual context.

\vspace{-2pt}
\paragraph{Evolution in individual cases.}
\Cref{fig:asynchronous-evidence} complements aggregate errors with observed
CXR sequences, representation forecasts, and separate clinical rollouts.
The examples include declining measurements and late changes missed by rollouts.

\begin{figure}[!t]
\centering
\includegraphics[width=\textwidth]{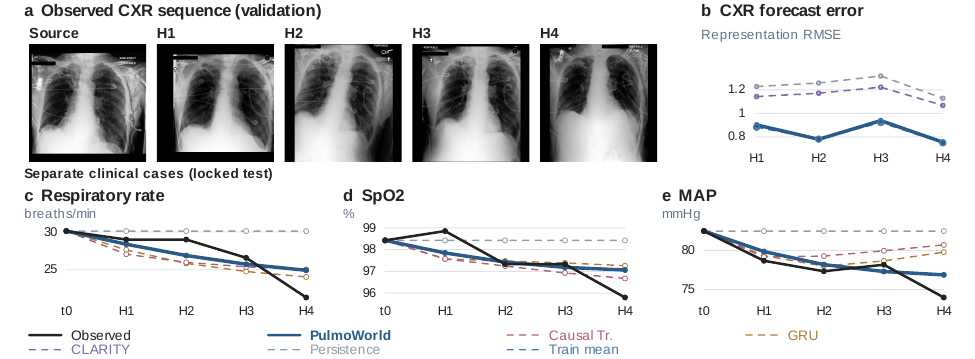}
\caption{\textbf{Observed imaging evolution and recursive clinical forecasts.}
(a) Observed MIMIC validation CXRs, not generated images or future rollout inputs.
H1--H4 denote six-hour windows, not acquisition times.
(b) Representation RMSE for the same sequence.
(c--e) Separate, outcome-selected locked-test clinical rollouts.
Selection and image timing are in
\cref{app:integrated-trajectories,fig:same-patient-io}.}
\label{fig:asynchronous-evidence}
\vspace{-3pt}
\end{figure}

\begin{samepage}
\paragraph{Reconstruction-to-rollout training.}
A matched MIMIC development study replaced the first three of twelve
rollout-training epochs with reconstruction, keeping the optimizer-step
budget fixed. The paired RMSE reduction was $0.00371$ [$0.00196,0.00535$]
(\cref{tab:recon-curriculum}).\par
\end{samepage}

\vspace{-2pt}
\paragraph{Information in simulated future states.}
One frozen outcome probe per cohort tests whether simulated future states
retain prognostic information about in-hospital death on MIMIC and 28-day death on MASS.
Compared with current-state inputs, future-state
summaries give AUROC $0.7135$ versus $0.7106$ on MIMIC and $0.7186$ versus $0.6856$ on MASS; AUPRC is $0.3237$ versus $0.3022$ and $0.5468$ versus $0.4263$, respectively.
MIMIC uses these 300 pneumonia test patients at H4; MASS uses a 99-patient validation panel
at available H2--H4 endpoints (\cref{tab:t4-full}).
The shared probe scores source observations together with endpoint changes.

\input{tables/tab_t4_main}

\vspace{-2pt}
\paragraph{Conditioning on recorded care.}
\label{sec:action-results}
Logged fluid--vasopressor inputs reduced MIMIC validation RMSE by $0.61\%$
[$0.45,0.80$] on action-exposed windows and $0.20\%$ overall.
Four train-derived care templates produce distinct MAP, lactate, and urine-output
trajectories from the same state (\cref{fig:action-planning-summary}a--c).
Selection among those continuations is then tested in MASS-Syn.

\vspace{-3pt}
\subsection{Evidence-constrained planning}
\label{sec:planning-results}

In five known-mechanism MASS-Syn environments, the fallback, the soft-penalty
policy, and Cap-MBRL share one authorized action set
(\cref{tab:planning-core}).
That set is one action-channel certificate and stays fixed at every simulated state. Ungated controls still select actions outside that authorized set. The unauthorized-action rate in \cref{tab:planning-core} counts executed actions outside the permitted set, so a zero rate means the mask was respected. Selection regret is the gap to the best authorized action. Cap-MBRL applies the risk budget and the predicted-cost margin, and this replacement of direct fallback execution moves regret from $0.091$ to $0.018$. The authorized soft-penalty policy is a separate objective inside the same set and remains at $0.3636$ (\cref{fig:action-planning-summary}d,e). The upper rows of \cref{tab:planning-core} execute outside that certificate, so their selection-regret cells are blank. The lower block keeps unauthorized rates at zero, with regrets $0.3636$, $0.0910$, and $0.0180$.

\input{tables/tab_planning_core}

\FloatBarrier
\begin{figure}[H]
\centering
\includegraphics[width=\textwidth]{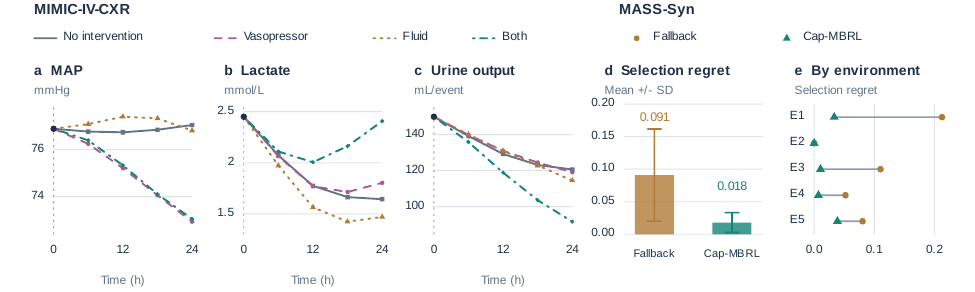}
\caption{\textbf{Conditional trajectories and authorized-set selection.}
(a--c) Conditional MIMIC validation forecasts under four six-hour care templates;
zero fluid and vasopressor inputs denote ``No intervention'' (\cref{app:action-visuals}).
(d,e) MASS-Syn selection regret: five-environment mean $\pm$ population SD
and paired environment results.}
\label{fig:action-planning-summary}
\vspace{-3pt}
\end{figure}
\endgroup

%% file: tables/tab_history_retention.tex
\begin{table}[!htbp]
\centering
\caption{\textbf{Asynchronous history filtering improves prognostic prediction.}
MIMIC development, three-seed means on 3,758 patients (524 deaths).
The asynchronous filter retains ordered history; fusion uses latest values.
Bold marks the better configuration within each reconstruction setting.}
\label{tab:history-retention}
\footnotesize
\setlength{\tabcolsep}{5pt}
\renewcommand{\arraystretch}{1.05}
\begin{tabular*}{\textwidth}{@{\extracolsep{\fill}}lccccc@{}}
\toprule
State estimation & AUROC $\uparrow$ & AUPRC $\uparrow$ & ACC $\uparrow$ & Bal. ACC $\uparrow$ & Brier $\downarrow$\\
\midrule
\multicolumn{6}{@{}l}{\textit{Without auxiliary reconstruction}}\\
Asynchronous filtering & \textbf{0.7565} & \textbf{0.3737} & \textbf{0.8481} & \textbf{0.6092} & \textbf{0.1064}\\
Latest-value fusion & 0.7352 & 0.3538 & 0.8454 & 0.6061 & 0.1079\\
\addlinespace[3pt]
\multicolumn{6}{@{}l}{\textit{With auxiliary reconstruction}}\\
Asynchronous filtering & \textbf{0.7748} & \textbf{0.4326} & \textbf{0.8673} & \textbf{0.6311} & \textbf{0.0965}\\
Latest-value fusion & 0.7367 & 0.3548 & 0.8427 & 0.6082 & 0.1075\\
\bottomrule
\end{tabular*}
\par\smallskip
\begin{minipage}{\textwidth}\footnotesize
ACC uses the frozen probe threshold of $0.23$; Bal.\ ACC averages sensitivity
and specificity. Majority-class ACC is $0.8606$, so thresholded accuracy is
descriptive, not the primary evidence of prognostic utility.
\end{minipage}
\end{table}

%% file: tables/tab_t2_main.tex
\begin{table}[!htbp]
\centering
\caption{\textbf{MIMIC-IV-CXR locked-test longitudinal simulation.}
H1--H4: macro NMAE; $\Delta$-NMAE: consecutive-change error with patient-bootstrap
$95\%$ CI; H4 dir.: direction agreement. Bold/underline denote best and second-best task-trained result in each column.
}
\label{tab:t2-main}
\scriptsize
\setlength{\tabcolsep}{2.4pt}
\renewcommand{\arraystretch}{0.82}

{\footnotesize\textbf{MIMIC-IV-CXR pneumonia locked test (300 patients)}}\\[2pt]
\resizebox{\textwidth}{!}{%
\begin{tabular}{@{}lccccccc@{}}
\toprule
\rowcolor{tblhead}
Method & H1$\downarrow$ & H2$\downarrow$ & H3$\downarrow$ & H4$\downarrow$ & H4 EHR-S@15$\uparrow$ & $\Delta$-NMAE$\downarrow$ & H4 dir.$\uparrow$ \\
\midrule
\rowcolor{tblhead!45}
\multicolumn{8}{@{}l}{\textbf{Controls}} \\
Persistence & 0.6740 & 0.7491 & 0.7692 & 0.7752 & 49.4\% & 0.443 [0.425,0.461] & 56.8\% \\
Train median & 0.7188 & 0.7009 & 0.6839 & 0.7090 & 48.8\% & 0.471 [0.454,0.489] & 59.6\% \\
\addlinespace[0pt]
\rowcolor{tblhead!45}
\multicolumn{8}{@{}l}{\textbf{General-purpose Direct VLMs}} \\
Qwen2.5-VL-3B$^{\dagger}$ & 0.6740 & 0.7491 & 0.7692 & 0.7752 & 49.4\% & 0.443 [0.425,0.461] & 56.8\% \\
Qwen2.5-VL-7B & 2.0176 & 2.8403 & 2.1843 & 1.8233 & 44.7\% & 0.682 [0.572,0.795] & 56.7\% \\
Qwen2.5-VL-72B & 0.9069 & 2.2291 & 0.8848 & 3.0042 & 44.0\% & 0.774 [0.446,1.410] & 57.0\% \\
Qwen3-VL-8B-Instruct & 4.5230 & 5.5157 & 6.1501 & 6.2845 & 46.0\% & 0.467 [0.449,0.486] & 57.0\% \\
Qwen3-VL-30B-A3B & 33.2483 & 35.0294 & 40.3677 & 25.2325 & 45.2\% & 0.581 [0.469,0.754] & 56.1\% \\
InternVL3-8B$^{\dagger}$ & 0.6740 & 0.7491 & 0.7692 & 0.7752 & 49.4\% & 0.443 [0.425,0.461] & 56.8\% \\
LLaVA-OV-1.5-8B & 2.4461 & 1.0740 & 3.4511 & 1.2832 & 38.3\% & 0.644 [0.564,0.742] & 55.5\% \\
\addlinespace[0pt]
\rowcolor{tblhead!45}
\multicolumn{8}{@{}l}{\textbf{Medical Direct VLMs}} \\
Lingshu-7B & 1.0245 & 1.1025 & 1.1310 & 1.1103 & 46.9\% & 0.458 [0.430,0.500] & 56.8\% \\
Hulu-Med-7B & 24.1641 & 11.6647 & 14.1120 & 9.8788 & 47.0\% & 0.897 [0.564,1.500] & 57.7\% \\
Hulu-Med-32B & 54.5972 & 34.4492 & 45.6867 & 22.0711 & 47.8\% & 0.588 [0.447,0.845] & 57.6\% \\
\addlinespace[0pt]
\rowcolor{tblhead!45}
\multicolumn{8}{@{}l}{\textbf{Task-trained models: temporal, adapted VLM, and world model}} \\
GRU & 0.5068 & \textbf{0.5483} & \underline{0.5818} & \underline{0.5934} & 52.1\% & 0.412 [0.396,0.428] & 60.4\% \\
Causal Transformer & \textbf{0.5005} & 0.5527 & 0.5877 & 0.5954 & \underline{52.5\%} & \underline{0.411 [0.395,0.428]} & 60.1\% \\
GRU-D & 0.5238 & 0.6815 & 0.8418 & 0.9600 & 38.3\% & 0.444 [0.429,0.461] & 58.1\% \\
Qwen2.5-VL-7B SFT & 0.6484 & 1.6620 & 0.7182 & 0.7221 & 49.5\% & 0.875 [0.589,1.415] & 56.4\% \\
CLARITY (adapted)$^{\P}$ & 0.5110 & 0.5746 & 0.6129 & 0.6328 & 50.8\% & 0.531 [0.513,0.550] & \underline{61.4\%} \\
\addlinespace[0pt]
\rowcolor{tblhead!45}
\multicolumn{8}{@{}l}{\textbf{Ours}} \\
\rowcolor{tblours}
\method{} & \underline{0.5034} & \underline{0.5489} & \textbf{0.5722} & \textbf{0.5780} & \textbf{54.9\%} & \textbf{0.407 [0.392,0.423]} & \textbf{61.7\%} \\
\bottomrule
\end{tabular}}

\vspace{2pt}
\parbox{0.98\textwidth}{\footnotesize
$^{\P}$Independently adapted CLARITY. $^{\dagger}$This Direct VLM row repeats the persistence control on every cell.
EHR-S@15: feature-macro success within 15\% relative error, with denominator
$\max(|y|,0.1\operatorname{IQR}_{\mathrm{train}})$; a compatibility score.
Detailed handling: Appendix~\ref{app:locked-forecast}.}
\end{table}

%% file: tables/tab_t4_main.tex
\begin{table}[!htbp]
\centering
\caption{\textbf{Cross-dataset prognosis metrics from predicted future states.}
One frozen probe per cohort: MIMIC pneumonia locked test ($n=300$) and MASS frozen
validation ($n=99$). MIMIC uses H4; MASS uses available endpoints through H4.
ACC uses frozen calibration-selected thresholds.
Bold and underline mark the best and second-best task-trained result in each column.}
\label{tab:t4-full}
\scriptsize
\setlength{\tabcolsep}{2.6pt}
\renewcommand{\arraystretch}{1.00}

\begin{tabular}{@{}lcccc@{\hspace{6pt}}cccc@{}}
\toprule
& \multicolumn{4}{c}{\cellcolor{tblmimic}\textbf{MIMIC locked test: in-hospital death}}
& \multicolumn{4}{c}{\cellcolor{tblmass}\textbf{MASS frozen validation: 28-day death}} \\
\cmidrule(lr){2-5}\cmidrule(lr){6-9}
\rowcolor{tblhead}
Method & ACC$\uparrow$ & AUROC$\uparrow$ & AUPRC$\uparrow$ & Brier$\downarrow$
& ACC$\uparrow$ & AUROC$\uparrow$ & AUPRC$\uparrow$ & Brier$\downarrow$ \\
\midrule
\rowcolor{tblhead!45}
\multicolumn{9}{@{}l}{\textbf{Controls}} \\
Current state & 0.8400 & 0.7106 & 0.3022 & 0.0899 & 0.6162 & 0.6856 & 0.4263 & 0.2051 \\
Persistence & 0.5500 & 0.6244 & 0.2139 & 0.1299 & 0.5960 & 0.6841 & 0.4254 & 0.2100 \\
Train median & 0.4200 & 0.5635 & 0.1947 & 0.1220 & 0.6263 & 0.6408 & 0.3776 & 0.2070 \\
\addlinespace[1pt]
\rowcolor{tblhead!45}
\multicolumn{9}{@{}l}{\textbf{General-purpose Direct VLMs}} \\
Qwen2.5-VL-3B$^{\dagger}$ & 0.8400 & 0.7106 & 0.3022 & 0.0899 & 0.6162 & 0.6856 & 0.4263 & 0.2051 \\
Qwen2.5-VL-7B & 0.6000 & 0.6179 & 0.1740 & 0.1765 & 0.6162 & 0.6851 & 0.4237 & 0.2052 \\
Qwen2.5-VL-72B & 0.6700 & 0.6817 & 0.2457 & 0.1229 & 0.6162 & 0.6987 & 0.4463 & 0.2056 \\
Qwen3-VL-8B-Instruct & 0.6133 & 0.6343 & 0.1597 & 0.1778 & 0.5960 & 0.6585 & 0.3613 & 0.2425 \\
Qwen3-VL-30B-A3B & 0.7400 & 0.6032 & 0.2429 & 0.1214 & 0.6061 & 0.6796 & 0.4149 & 0.2129 \\
InternVL3-8B$^{\dagger}$ & 0.8400 & 0.7106 & 0.3022 & 0.0899 & 0.6162 & 0.6856 & 0.4263 & 0.2051 \\
LLaVA-OV-1.5-8B & 0.4667 & 0.6249 & 0.1988 & 0.2069 & 0.6566 & 0.7057 & 0.4471 & 0.2088 \\
\addlinespace[1pt]
\rowcolor{tblhead!45}
\multicolumn{9}{@{}l}{\textbf{Medical Direct VLMs}} \\
Lingshu-7B$^{\ddagger}$ & 0.8400 & 0.7106 & 0.3022 & 0.0899 & 0.6364 & 0.6650 & 0.3822 & 0.2129 \\
Hulu-Med-7B & 0.7567 & 0.6582 & 0.1651 & 0.1631 & 0.5960 & 0.6730 & 0.4216 & 0.2073 \\
Hulu-Med-32B & 0.7600 & 0.6587 & 0.2294 & 0.1113 & 0.6364 & 0.5986 & 0.3790 & 0.2368 \\
\addlinespace[1pt]
\rowcolor{tblhead!45}
\multicolumn{9}{@{}l}{\textbf{Task-trained models: temporal, adapted VLM, and world model}} \\
GRU & 0.5767 & 0.5977 & 0.2381 & 0.1201 & 0.6229 & 0.6442 & 0.3931 & 0.2385 \\
Causal Transformer & 0.5967 & 0.6059 & 0.2494 & 0.1147 & 0.5993 & 0.6524 & 0.3992 & 0.2439 \\
GRU-D & 0.6200 & 0.6382 & 0.2285 & 0.1342 & 0.6061 & 0.6715 & 0.3828 & 0.2146 \\
Qwen2.5-VL-7B SFT & \underline{0.8467} & \underline{0.7003} & \underline{0.2861} & \underline{0.0906} & 0.6162 & 0.6856 & 0.4263 & 0.2051 \\
CLARITY (adapted) & 0.5333 & 0.6040 & 0.2433 & 0.1273 & \underline{0.6465} & \underline{0.6932} & \underline{0.4636} & \underline{0.1907} \\
\addlinespace[1pt]
\rowcolor{tblhead!45}
\multicolumn{9}{@{}l}{\textbf{Ours}} \\
\rowcolor{tblours}
\method{} & \textbf{0.8500} & \textbf{0.7135} & \textbf{0.3237} & \textbf{0.0867} & \textbf{0.6768} & \textbf{0.7186} & \textbf{0.5468} & \textbf{0.1361} \\
\bottomrule
\end{tabular}

\vspace{1pt}
\parbox{0.98\textwidth}{\scriptsize
$^{\dagger}$This Direct VLM row repeats the current-state control on every cell. $^{\ddagger}$The MIMIC cells repeat the current-state control; the MASS cells do not. Cohort-specific fitting details: Appendices~\ref{app:locked-forecast} and~\ref{app:vlm-prognosis}.}
\end{table}

%% file: tables/tab_planning_core.tex
\begin{table}[!htbp]
\centering
\caption{\textbf{Policy controls for evidence-constrained planning.}
Five-environment MASS-Syn means (lower is better). Unauthorized-action rate is the
rate of executed actions outside the permitted set. Selection regret is the
gap to the best action inside that set. }
\label{tab:planning-core}
\footnotesize
\setlength{\tabcolsep}{4pt}
\renewcommand{\arraystretch}{0.94}
\begin{tabular}{@{}lcc@{}}
\toprule
Policy control & Unauth. (\%) & Sel.\ regret \\
\midrule
Unrestricted WM-RL & 32.19 & --\\
Uncertainty-aware & 32.26 & --\\
Support-aware & 33.71 & --\\
Path-consistency weighting & 31.58 & --\\
\midrule
\rowcolor{tblhead}
\multicolumn{3}{l}{\textbf{Same authorized action set}}\\
Authorized + soft penalties & 0.00 & 0.3636 \\
Authorized policy (fallback) & 0.00 & 0.0910 \\
\rowcolor{tblours}
Cap-MBRL & 0.00 & \textbf{0.0180} \\
\bottomrule
\end{tabular}
\end{table}

%% file: sections/07_limitations.tex
\vspace{-2pt}
\section{Conclusion}
\method{} is a world model for longitudinal pneumonia dynamics. It keeps
incomplete clinical, imaging, and BALF evidence in one belief, rolls that
belief forward, and compares actions only inside an authorized set. The two
cohorts are trained and evaluated separately. On MASS, recursive forecasts
are competitive with task-trained sequence models, and simulated future states
retain prognostic information about 28-day death. On MIMIC-IV-CXR, history-aware
filtering improves prognostic discrimination over latest-value fusion in
development. Locked evaluation uses 300 pneumonia patients drawn from the
1,178 pneumonia cases in the 6,535-patient test split. On that panel, the forecaster is competitive with task-trained models on long-horizon error, and simulated states retain information
about in-hospital death. In five known-mechanism MASS-Syn environments,
Cap-MBRL lowers selection regret from $0.091$ under direct execution of the
frozen fallback to $0.018$.

%% file: appendix/00_roadmap.tex
\section*{Appendix Contents}
\phantomsection
\addcontentsline{toc}{section}{Appendix Contents}

The supplementary material develops the evidence behind history-aware state
estimation, multimodal forecasting, and evidence-constrained action selection.
Sections A and B define the data and models; Section C presents full
comparisons and component tests; Section D connects predictions to observed
patient records; Section E specifies reproducibility and access conditions.
Locked-test results, development studies, and synthetic experiments are
identified separately throughout. Titles and page numbers below are linked.

\begin{center}
\small\renewcommand{\arraystretch}{1.28}
\begin{tabular}{@{}p{0.08\textwidth}p{0.78\textwidth}r@{}}
\toprule
 & Contents & Page \\
\midrule
A & \hyperref[app:data]{\textbf{Data construction, distributions, and privacy}} & \pageref{app:data} \\
 & \hyperref[app:data-access]{Public MIMIC, private MASS, and synthetic MASS-Syn} & \pageref{app:data-access} \\
 & \hyperref[app:mass-construction]{MASS event construction and evaluation populations} & \pageref{app:mass-construction} \\
 & \hyperref[app:mimic-construction]{MIMIC alignment, aggregation, and patient splits} & \pageref{app:mimic-construction} \\
 & \hyperref[app:distributions]{Patient proportions, modality coverage, event/window counts} & \pageref{app:distributions} \\
 & \hyperref[app:missingness]{Observed-value masks, reconstruction targets, and leakage control} & \pageref{app:missingness} \\
 & \hyperref[app:ethics]{Data permissions and privacy} & \pageref{app:ethics} \\
\addlinespace
B & \hyperref[app:implementation]{\textbf{Model specification, training, and evaluation}} & \pageref{app:implementation} \\
 & \hyperref[app:architecture]{Model families, tensor dimensions, parameter counts} & \pageref{app:architecture} \\
 & \hyperref[app:training]{Training phases and checkpoint selection} & \pageref{app:training} \\
 & \hyperref[app:method-details]{Belief updates, authorization, and full policy equations} & \pageref{app:method-details} \\
 & \hyperref[app:execution]{Dataset-specific execution and action interfaces} & \pageref{app:execution} \\
 & \hyperref[app:prognosis-implementation]{Prognosis comparator and outcome probe} & \pageref{app:prognosis-implementation} \\
 & \hyperref[app:planner-implementation]{Policy fitting and decision-time search} & \pageref{app:planner-implementation} \\
\addlinespace
C & \hyperref[app:results]{\textbf{Complete quantitative results}} & \pageref{app:results} \\
 & \hyperref[app:locked-forecast]{Full MASS and MIMIC locked-test forecasting tables} & \pageref{app:locked-forecast} \\
 & \hyperref[app:extended-rollout-evidence]{Future-state prognosis and representation forecasts} & \pageref{app:extended-rollout-evidence} \\
 & \hyperref[app:t1-state]{Masked source reconstruction (T1)} & \pageref{app:t1-state} \\
 & \hyperref[app:mechanism-evidence]{Component ablations and missing-evidence sensitivity} & \pageref{app:mechanism-evidence} \\
 & \hyperref[app:mimic-history-prog]{History-enabled MIMIC prognosis follow-up} & \pageref{app:mimic-history-prog} \\
 & \hyperref[app:vlm-prognosis]{Task-adapted VLM prognosis evaluation} & \pageref{app:vlm-prognosis} \\
 & \hyperref[app:mimic-pred-action]{MIMIC action conditioning and candidate ranking} & \pageref{app:mimic-pred-action} \\
 & \hyperref[app:cgrpi]{Known-mechanism policy selection and gate diagnostics} & \pageref{app:cgrpi} \\
 & \hyperref[app:action-visuals]{Action-conditioned branches and authorization evidence} & \pageref{app:action-visuals} \\
\addlinespace
D & \hyperref[app:raw-cases]{\textbf{Case atlas: inputs, future states, and failure modes}} & \pageref{app:raw-cases} \\
 & \hyperref[app:matched-records]{Matched source/next-state records and complete input fields} & \pageref{app:matched-records} \\
 & \hyperref[app:worked-outputs]{Same-patient multimodal inputs, complete outputs, and masks} & \pageref{app:worked-outputs} \\
 & \hyperref[app:diverse-cases]{Diverse CT pairs and CXR/action input examples} & \pageref{app:diverse-cases} \\
 & \hyperref[app:integrated-trajectories]{Clinical case selection and image-state comparison} & \pageref{app:integrated-trajectories} \\
 & \hyperref[app:frozen-representation]{Longitudinal imaging and metabolomic evolution} & \pageref{app:frozen-representation} \\
\addlinespace
E & \hyperref[app:reproducibility]{\textbf{Reproducibility and data access}} & \pageref{app:reproducibility} \\
\bottomrule
\end{tabular}
\end{center}
\clearpage

%% file: appendix/a_data_privacy.tex
\section{Data Construction, Splits, Ethics, and Privacy}
\label{app:data}

\subsection{Public MIMIC, private MASS, and synthetic selection}
\label{app:data-access}
The paper uses two real cohorts and one simulator. They differ in access,
clinical scope, and which claim they can support.

MIMIC-IV-CXR is assembled from public releases: MIMIC-IV clinical records
\citep{johnson2023mimic} and MIMIC-CXR chest radiographs
\citep{johnson2019mimiccxr}. Both are available from PhysioNet under a
credentialed data-use agreement. We do not redistribute the source files.
The linked records supply chest radiographs and clinical trajectories for
training and evaluation. They support recursive forecasting, history
retention, and logged fluid--vasopressor conditioning.

MASS is a private institutional cohort. It links pneumonia-related clinical
measurements, chest CT, and BALF metabolomics. Patient-level records are
not publicly released. MASS supports the pneumonia-specific representation
and locked Clinical19 forecast. External repetition of those patient-level
analyses requires controlled access, which is separate from the public code
and aggregate-metric release (\cref{app:reproducibility}).

MASS-Syn contains no patient records. It is a known-mechanism simulator and
is used only to score authorized-set selection.

\begin{figure*}[t]
  \centering
  \includegraphics[width=\textwidth]{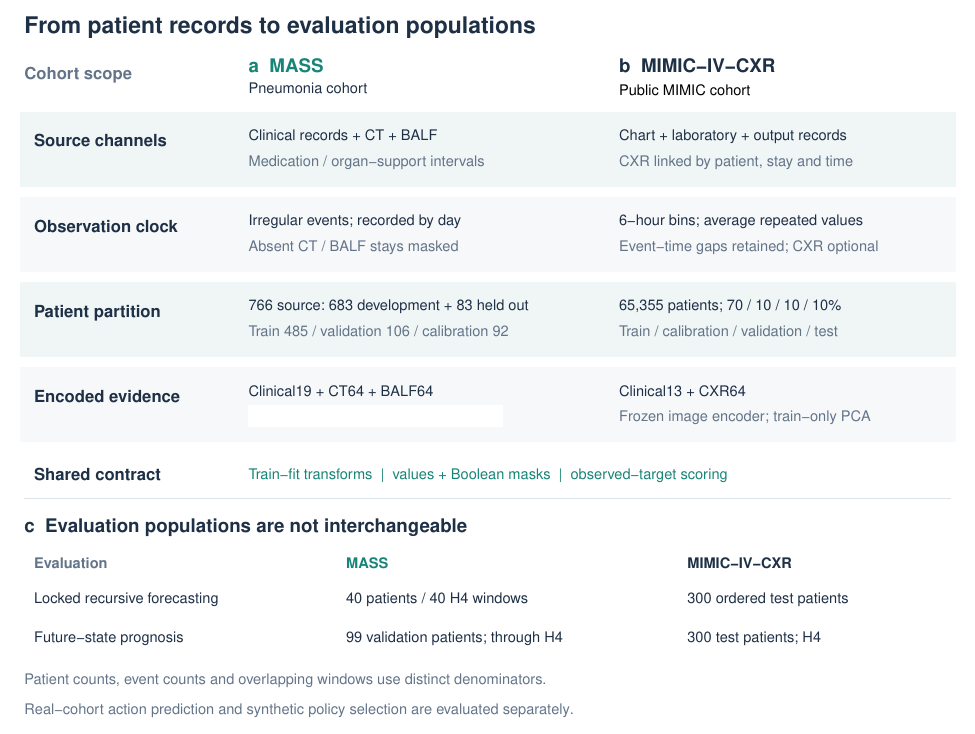}
  \caption{\textbf{From cohort records to model-specific evaluation populations.}
  (a,b) Private MASS and public MIMIC differ in clinical scope, observation
  clock, and modality availability. Patient splits precede model fitting; numerical
  transforms and projections are train-fit. (c) Forecasting and prognosis
  use distinct eligibility rules and must retain their own denominators.
  MASS prognosis uses the available endpoint through H4, not uniformly H4.
  Counts describe frozen data extracts, not recruitment estimates.}
  \label{fig:data-construction}
\end{figure*}

\begin{figure*}[t]
\centering
\includegraphics[width=\textwidth]{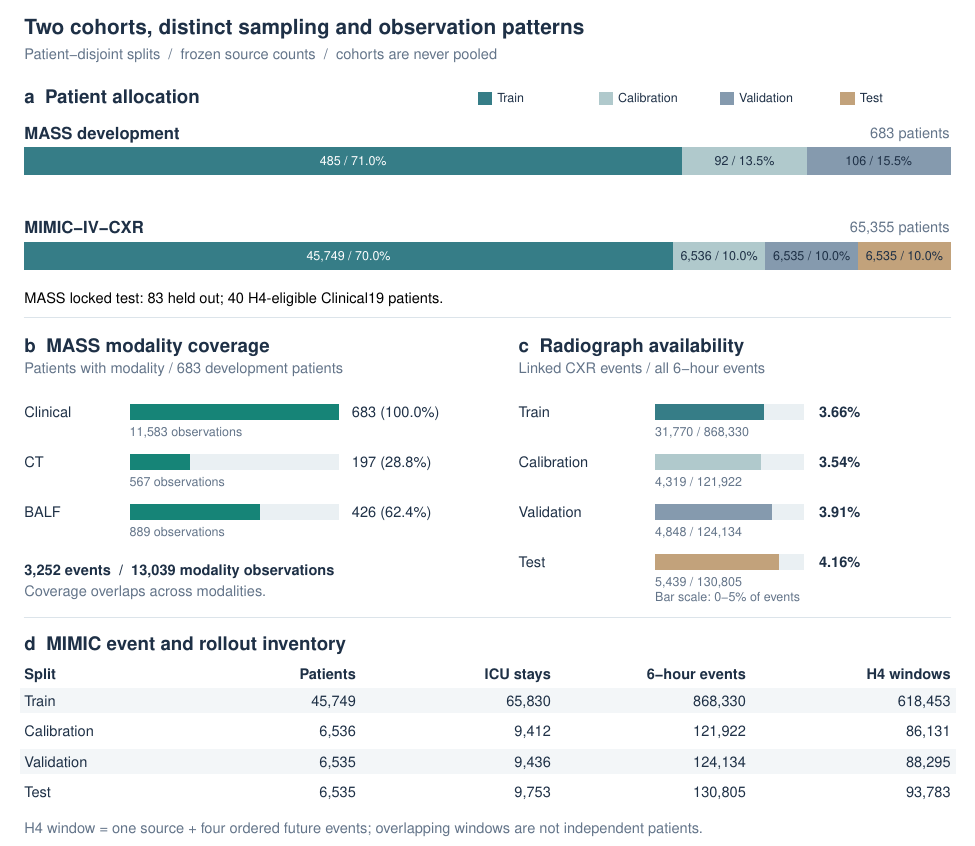}
\caption{\textbf{Cohort composition and observation availability.}
(a) Patient counts and within-cohort split proportions; MASS development
excludes the 83 locked-test patients. The Clinical19 recursive forecasting panel contains 40
H4-eligible patients and 40 evaluated windows.
(b) MASS patient-level modality coverage, with observation counts shown
separately. Its 567 CT observations occur on 3,252 irregular events (17.4\%).
(c) MIMIC linked-radiograph counts divided by all 6-hour events in each split.
Patient coverage in (b) and event availability in (c) have different
denominators. (d) Exact MIMIC patient, stay, event, and H4-window counts.
An H4 window contains one source and four ordered future events; overlapping
windows are not independent patients.}
\label{fig:data-distribution}
\end{figure*}

Throughout this appendix we keep three counting units separate: the source
cohort, the model-specific data view, and the eligible evaluation population.
One patient can contribute multiple events and forecast windows, so those
record counts are not independent sample sizes.
Figure~\ref{fig:data-construction} maps construction choices to evaluation
populations. Figure~\ref{fig:data-distribution} reports split proportions
and observation coverage.

\subsection{MASS construction and evaluation populations}
\label{app:mass-construction}
MASS is the private pneumonia cohort (\cref{app:data-access}). It begins
with patient-linked clinical records, CT studies, BALF
metabolomic profiles, and timestamped medication or organ-support intervals.
Clinical values and their observation masks are aligned to the irregular
event clock at recorded day-level resolution. Adjacent events form factual
transitions. Interval actions
describe exposure between source and target. A missing CT or BALF record
does not create a synthetic observation. Clinical19 is the interface for
development and for the 40-patient locked forecast. The counts below
are this Clinical19 cohort.

\paragraph{Source linkage and record eligibility.}
The recorded clinical source contains 766 patients. Patient identifiers
link modalities before the 83-patient locked partition is excluded,
leaving 683 development patients. Clinical, CT, and BALF records need
not coincide in time or be available for every patient.
The 101 omics records lacking verified enrollment linkage are omitted
at the record level, not counted as 101 patient exclusions.
These counts describe the available source extract, not an institutional
recruitment or diagnostic-screening denominator.

\paragraph{Clinical19 evaluation populations.}
The development extract contains 3,252 events and 2,569 adjacent transitions;
the latter split into 1,806/428/335 train/validation/calibration transitions.
Clinical19 state rows on this split number 2,608. Eligible rollout populations
contain 465/104/85 patients and 1,971/461/368 windows in the same split order.
The locked Clinical19 comparison uses 40 eligible patients and
40 four-step windows from the separate held-out population.

\subsection{MIMIC-IV-CXR construction and temporal alignment}
\label{app:mimic-construction}
This section describes our alignment of the public MIMIC-IV and MIMIC-CXR
releases (\cref{app:data-access}). The 65,355-patient extract supplies the trajectories used for training and
evaluation, including stays without a linked radiograph.
Eligible chart, laboratory, and output records are linked to ICU stays.
Implausible values are excluded by the frozen variable rules. Measurements
are assigned to 6-hour bins relative to ICU admission, and repeated eligible
values of a feature within a bin are averaged. Radiographs are linked to the
same patient, stay, and time bin. The union of clinical and radiograph bins
defines the event table; masks retain which features and images were observed.
H4 examples contain one source and four ordered future events from one stay.
Elapsed gaps come from event times rather than fabricated intermediate states.

Final patient allocation targets 70/10/10/10\% for
train/calibration/validation/test. Official CXR validation and test patients
retain those holdouts; remaining patients are assigned by a deterministic
SHA-256 order to fill the quotas. This world-model partition is distinct
from simply using the original image-file split. The locked recursive forecasting comparison
uses 300 pneumonia patients drawn from the 1,178 pneumonia cases in the
6,535-patient test partition, after freezing the evaluation protocol.

\paragraph{Forecast windows and care intervals.}
A forecast window is eligible only within one ICU stay, with a source event
and four ordered future events. A six-hour bin width does not require
consecutive observed events to be exactly six hours apart. Clinical targets
are scored only where their feature masks indicate an observed value.
Linked CXR availability is retained separately from clinical completeness;
patients without a radiograph remain in the clinical cohort.
Fluid and vasopressor exposures describe source-to-target care intervals,
rather than future clinical measurements supplied to the forecast.

\subsection{Observation and split distributions}
\label{app:distributions}

\paragraph{Evaluation interfaces.}
Present-state reconstruction recovers masked observed values; recursive
forecasting predicts H1--H4 without factual intermediate observations.
Future-state prognosis applies a frozen cohort-specific outcome probe.
Action evaluation separates logged-care prediction from authorization
and known-mechanism policy selection. MASS uses Clinical19. MIMIC uses 13 clinical variables and CXR.
The two cohorts are not pooled.

\paragraph{Development population and eligible modeling records.}
The counts below itemize the private MASS development extract already
defined in \cref{app:mass-construction}.
The test-excluded development view contains 683 patients, 3,252 timestamped
events, 13,039 modality observations, 3,409 interval action records, and 2,569
adjacent factual transitions. Patient-level train/validation/calibration sizes
are 485/106/92. The 83 locked-test patients are physically excluded from development
evaluation. Clinical, CT, and BALF source observations cover 683/197/426
patients; 101 omics records without verified enrollment mapping are omitted.

The Clinical19 development model re-extracts clinical state from the same
patient-disjoint cohort. The frozen state table contains 2,608 event rows
and 23,724 observed feature rows: 1,853/413/342 state rows and
16,833/3,751/3,140 observations for train/validation/calibration. Nineteen
clinical features are used; ten have eligible repeated-measurement targets
and nine are baseline context. Grouping, units, transforms, and eligibility
are frozen from train and calibration before validation.

\input{tables/tab_cohort}

MIMIC action construction aligns 1,150,760 adjacent transitions. Train-only
cut points define four fluid and four vasopressor levels, yielding 16 supported
cells (minimum train support 200). The no-recorded-action cell contains 607,910
of 802,500 train transitions (75.8\%). Source extraction retains 163,213
qualifying crystalloid events and 735,644 convertible vasopressor events;
excluded or unaligned records are logged rather than silently assigned.

CT embeddings are computed from a representative 2D slice by a frozen
Qwen2.5-VL-7B and reduced from 3,584 to 64 dimensions by train-only PCA. BALF
metabolomics is represented by a train-only 64-dimensional projection. No CT
pixel is decoded. The source action ontology maps 137 medication names and
organ-support records into 20 macro classes. The MASS world model evaluated
in this paper keeps this ontology for auditing and disables its care-context
residual. Releasing that ontology does not release the private records.

\subsection{Missingness, transforms, and leakage control}
\label{app:missingness}
Clinical transforms subtract the train median and divide by the train
interquartile range, with a unit fallback only for zero or non-finite IQR.
Missing clinical entries use the train median solely as a finite storage
value, yielding zero after normalization while the Boolean mask remains false.
The encoder receives both zero-masked values and feature-level masks, so an
observed median and an unobserved value remain distinguishable.
Entirely absent modalities have zero correction gain; elapsed-time dynamics
still advance the prior state. Observation-value losses and evaluation use
only observed targets, whereas the availability head is supervised by both
observed and missing target masks.
Missing CT or BALF remains masked rather than synthesized. Versioned
physiologic bounds quarantine out-of-range records rather than repair them.

\paragraph{Observed, absent, and deliberately hidden values.}
An observed median has normalized value zero and mask one; a missing value
has the same storage value but mask zero. Only the former supplies an
observed-value target. During source-reconstruction pretraining, a subset
of observed entries is deliberately hidden from the encoder while its
original values remain reconstruction targets. This artificial training
mask is distinct from natural missingness. At inference, missing-evidence
perturbations remove available inputs without converting genuinely absent
values into targets (\cref{tab:missing-evidence}). Thus the mask specifies
which evidence is available, not merely a numerical imputation convention.

Preprocessing statistics, projection parameters, action-cell definitions,
propensity models, and 28-day outcome-probe hyperparameters are train-fit or
calibration-selected without test labels. The later Clinical19 and MIMIC
training phases use calibration for checkpoint selection and validation for
reporting. Locked MASS forecasts use those Clinical19 checkpoints.
Bootstrap resampling uses patients as units, retaining all observations from a
sampled patient.

\input{tables/tab_patient_flow}

\subsection{Data permissions and privacy}
\label{app:ethics}
MASS is a private institutional cohort used with permission for retrospective
research. Personal identifying information was removed from the research
data. MIMIC-IV and MIMIC-CXR are accessed under the PhysioNet data-use
agreement. These permissions do not authorize redistribution of patient-level
records.

The planned public release contains evaluation code, preprocessing specifications, synthetic
mechanisms, and aggregate metrics. It excludes patient-level MASS records
and does not repackage MIMIC source files. Released tables exclude direct or
hashed patient identifiers, dates, DICOM identifiers, and
free-text medication names. Image exemplars printed here remain under their
source agreements and are not a standalone public dataset.

%% file: tables/tab_cohort.tex
\begin{table}[t]
\centering
\caption{\textbf{Clinical19 development data view.} The 83-patient locked test is physically excluded; the 40-patient Clinical19 recursive forecasting evaluation is reported separately. Development splits are patient-disjoint.}
\label{tab:cohort}
\small
\begin{tabular}{@{}lr@{}}
\toprule
Quantity & Value \\
\midrule
Development patients & 683 \\
Patient split (train/validation/calibration) & 485 / 106 / 92 \\
Events / observations / factual transitions & 3,252 / 13,039 / 2,569 \\
Transition split (train/validation/calibration) & 1,806 / 428 / 335 \\
Patients with clinical / CT / BALF & 683 / 197 / 426 \\
Clinical / CT / BALF observations & 11,583 / 567 / 889 \\
Clinical / CT latent / metabolomics dimensions & 19 / 64 / 64 \\
Actions & 20 macro classes $\times$ 4 interval channels (80-D) \\
Retrospective / strict-online usable transitions & 2,569 / 285 \\
Locked-test patients included & 0 of 83 \\
\bottomrule
\end{tabular}
\end{table}

%% file: tables/tab_patient_flow.tex
\begin{table}[H]
\centering
\caption{\textbf{Patient flow into the development population.} Record-level quarantine rows are not unique-patient exclusions. Raw screening and institutional recruitment denominators remain unavailable.}
\label{tab:patient-flow}
\small
\begin{tabular}{@{}p{0.42\textwidth}rrp{0.22\textwidth}@{}}
\toprule
Flow step & Remaining & Removed & Evidence status \\
\midrule
Recorded clinical source cohort & 766 & -- & Source extract \\
Authoritative patient split assigned & 766 & 0 & Exact set match \\
Locked-test patients physically excluded & 683 & 83 & Test closed \\
Test-excluded development cohort & 683 & 0 & Exact non-test set \\
Patients with at least one event & 683 & 0 & Complete coverage \\
\bottomrule
\end{tabular}
\end{table}

%% file: appendix/b_implementation.tex
\FloatBarrier
\section{Model Specification, Training, and Evaluation Protocols}
\label{app:implementation}

MASS uses the Clinical19 checkpoint for development and locked forecasting.
MIMIC uses one history-aware setting: 13 clinical variables and CXR. The two
cohorts differ in dimensions and fitted parameters. Architecture and loss definitions appear first; execution,
prognosis, and planning details follow.

\input{appendix/b_architecture_details}

\input{appendix/b_method_details}

\subsection{Dataset-specific execution and action interfaces}
\label{app:execution}

The shared abstraction is a masked multimodal belief updated by elapsed time;
the concrete modules are sized separately for each dataset rather than sharing
a fitted forecasting head. \Cref{tab:model-implementation} reports the
configurations used for the quantitative and case analyses.

\begin{table*}[t]
\centering
\caption{\textbf{Model architecture and training hyperparameters.} Each cohort has one observation setting. Parameter
counts include learned world-model modules but exclude frozen image encoders
and train-fitted PCA. Constrained refinement continues from the Clinical19 checkpoint. The MIMIC action adapter initializes from the history-aware parent and updates only action residual and gate parameters.}
\label{tab:model-implementation}
\scriptsize
\setlength{\tabcolsep}{3pt}
\begin{tabular}{@{}p{0.18\textwidth}p{0.27\textwidth}p{0.24\textwidth}p{0.24\textwidth}@{}}
\toprule
Phase & Inputs and state & Optimization & Selection and size \\
\midrule
MASS multimodal & Clinical19; CT64; BALF64; seven 64-D compartments; hidden 256; 16 Fourier time features; bounded continuous time; action branch disabled. & Base prediction: AdamW, batch 16, learning rate $5\times10^{-5}$, at most 20 epochs. Constrained refinement: $10^{-5}$, at most 12 epochs, recursive/direct horizon weights 0.05/0.25/1.5/2.2. & Calibration selection; seeds 20260915--20260917; 5,338,172 learned parameters per checkpoint. \\
MIMIC & 13 clinical variables; CXR PCA64; two compartments (clinical/radiograph); hidden 64; compartment 32; 8 Fourier time features; four recursive horizons. & AdamW; batch 256; learning rate $10^{-3}$; weight decay $10^{-4}$; gradient clipping 5.0; 12 epochs; horizon weights 0.25/1/1.5/2. & Calibration recursive balanced RMSE; seeds 20260915--20260917; 192,461 learned parameters; action disabled. \\
MIMIC action-conditioned & Same parent plus two transformed factual action channels in a 6-D reserved interface; action residual and gate enabled. & AdamW on action adapter only; batch 256; learning rate $3\times10^{-4}$; weight decay $10^{-4}$; gradient clipping 1.0; at most 8 epochs; patience 2 after epoch 3. & Calibration clinical macro RMSE; selected epoch 7; 20,352 action-related parameters within the 192,461-parameter model. \\
\bottomrule
\end{tabular}
\end{table*}

For MIMIC, the action adapter is fit on the same history-aware parent. Its training uses 618,453 H4 windows; calibration and
validation contain 86,131 and 88,295 windows. 
Recursive loss, direct-horizon loss, availability loss, and horizon weights are
frozen in the configuration; the action adapter sets the direct-path
coefficient to zero and leaves the parent dynamics fixed.

Reconstruction and forecast errors are computed per patient, feature, and horizon before
aggregation. All methods on a cohort are scored on identical observed targets. Forecast
carry-forward is scored only when source and target are both observed.
Per-patient losses are averaged before paired bootstrap.

The action-information gate uses state-only and state-plus-action
predictors on a known-signal / null synthetic control (action coefficient
$0.8$ or $0$) and on eight real MASS cells. Acceptance requires a
prespecified MAE gain, a positive patient-bootstrap lower bound, factual
gains over global and propensity-matched shuffles, and at least $80\%$
propensity overlap. Both model families must pass. The action ontology is in \cref{tab:action-ontology}. The eight prespecified
real MASS cells do not open an authorized set, so planning is scored in MASS-Syn.

\input{tables/tab_action_ontology}

\subsection{Prognosis comparator and outcome probe}
\label{app:prognosis-implementation}
\paragraph{MASS CLARITY comparator.}
This arm is fitted independently to Clinical19 clinical variables and CT64, and does not receive BALF. It uses four Transformer layers,
hidden width 256, eight attention heads, dropout 0.1, batch size 64,
AdamW at $3\times10^{-4}$ with weight decay $10^{-4}$, gradient clipping 1,
and three training seeds. The 1,971 training pairs fit parameters; 368 calibration
pairs select among 40 epochs (selected epoch 6). Clinical and CT squared
errors have weight 1; availability cross-entropy has weight 0.1.
The unchanged shared probe evaluates the 99 validation patients, including
28 deaths. Clinical values and availability are updated recursively;
nonclinical probe coordinates remain at source values, matching the
Direct-VLM observable-interface scoring convention. 

The 28-day outcome probe is L2 logistic regression on train-standardized features.
It receives source observables together with endpoint-minus-source changes
and elapsed time, not the future state alone. The source summary includes
clinical values/masks and CT/BALF representation summaries.
The 99-patient MASS panel uses the available endpoint: H2, H3, and H4
for 2, 46, and 51 patients, respectively.
$C$, temperature, and threshold are selected on calibration. The matched
shuffle uses the same fitted probe.

\subsection{Policy fitting and decision-time search}
\label{app:planner-implementation}
When $G_A\cap G_I$ exposes $\mathcal A_{\rm auth}(B_t)$, we optimize a
two-layer state--action MLP $\pi_\phi$ on MASS-Syn only. Each hidden layer
has width $128$ and a $\tanh$ nonlinearity. The input is the $24$-dimensional
latent state plus a four-step indicator, and the output is seven action
logits ($21{,}127$ parameters;
\cref{fig:network-pipeline}). The state cost and
return are
\begin{equation}
c(B_h)=\sum_{k=1}^{3}w_k[(B^k_{h,1})^2+0.5(B^k_{h,2})^2],\quad
R(\tau)=-\sum_{h=1}^{H}\gamma^{h-1}c(B_h),
\end{equation}
with $w=(1,0.65,0.45)$. Model uncertainty, behavior surprisal,
recursive--direct discrepancy, and capability violation form a bounded risk
$\rho_h$. The no-action risk tightens
$b(B_t)=\operatorname{clip}(b_0-\alpha\rho_{0}(B_t),b_{\min},b_0)$.
Fitting uses the root-mean-square standardized group-relative advantage in
\cref{eq:cgrpi}, sampled KL to behavior, and a nonnegative dual
multiplier on mean budget violation. At inference a feasible action is
accepted only when it beats the authorized-set fallback by a
frozen margin.

\paragraph{What training compares.}
The behavior policy is trained by cross-entropy on the logged category.
Category 0 is the zero action. Categories 1--6 are the six one-hot channels.
That policy is then frozen. Actor training samples a category from the
authorized softmax and rolls it through the ensemble. Several samples that
share an initial state form a group. Each trajectory receives its discounted
negative severity cost minus an ensemble-uncertainty penalty. Its advantage
is the centered, standardized score within the group
(\cref{eq:group-advantage}); thus reinforcement rewards better predicted
trajectories without requiring oracle action labels.
The KL term penalizes departure from the behavior distribution. The
mean of the positive trajectory-level budget violations enters
\cref{eq:cgrpi} on the same categorical samples
(\cref{app:method-details}). Decision-time selection applies the same budget
to the action that is executed.

\paragraph{Inference and selection-regret evaluation.}
At each decision, the ranker enumerates candidate first actions and continues
each trajectory with the fitted policy through the frozen world-model ensemble.
A separate trajectory follows the authorized fallback. Candidate scores combine
discounted predicted state cost and bounded risk. The optional
policy-preference penalty has coefficient zero in the reported experiment.
The authorization mask and state-dependent risk budget restrict eligibility;
the fallback is retained unless a feasible candidate clears the score margin.
The selected first action is executed, and ranking repeats at the next state.
For evaluation only, the known simulator supplies each policy's realized
discounted cost and the lowest cost among authorized action sequences.
Their difference is selection regret; the oracle is not supplied to the planner.
Figure~\ref{fig:action-planning-summary}d compares this quantity across methods
with zero unauthorized action rate. It evaluates selection within the permitted set,
not the separate regret imposed by restricting that set.

%% file: appendix/b_architecture_details.tex
\subsection{Cohort-specific model families and tensor interfaces}
\label{app:architecture}

Observation modalities and internal state partitions are distinct.
MASS has three observation streams: structured clinical measurements,
CT representations, and BALF metabolomics. MIMIC-IV-CXR has two:
structured clinical measurements and CXR representations.
Both infer one patient state. The Clinical19 configuration groups its
clinical variables into respiratory (5), renal (3), inflammatory (4),
hemodynamic (4), and global-severity (3) blocks, with CT and BALF mapped
to imaging and metabolic blocks. These seven internal state blocks
parameterize the three observation modalities.
MIMIC instead uses two internal blocks, clinical and radiograph.

Locked MASS forecasting uses this Clinical19 checkpoint. It partitions 19
variables into groups of sizes 5/3/4/4/3 and maps CT64 and BALF64 onto the imaging and metabolic compartments. Each of these seven compartments stores a
64-dimensional mean and diagonal log variance. The MIMIC model has 13 clinical variables and radiograph64 inputs, two 32-dimensional compartments, and
hidden width 64. Its locked forecast uses the same history-aware filter that
the development comparison evaluates against latest-value fusion. Both
implementations use evidence-recency gating for cross-compartment messages.

\begin{figure}[t]
\centering
\includegraphics[width=\textwidth]{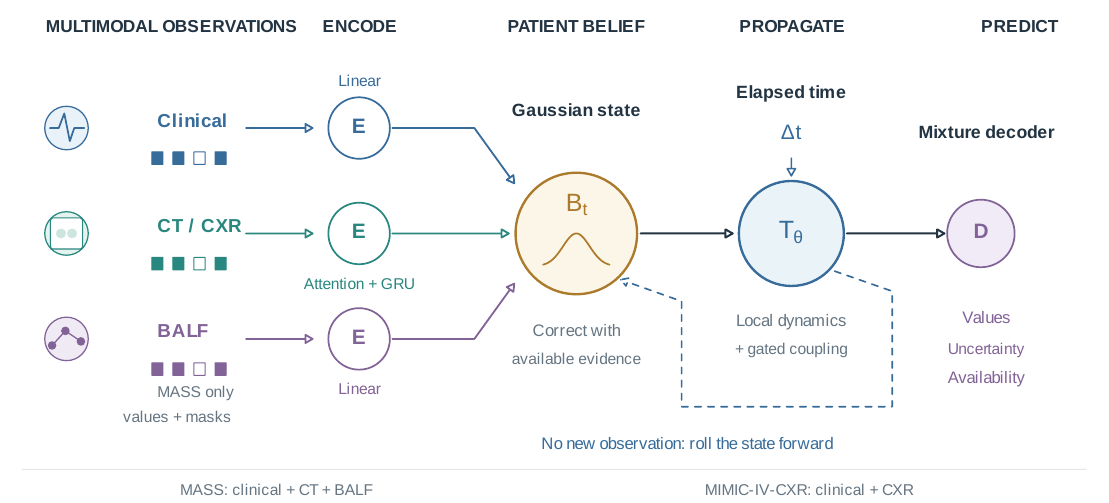}
\caption{\textbf{Architecture of the \method{} observation filter.}
Modality-specific encoders map masked observations into a Gaussian patient
belief; imaging features additionally pass through attention and recurrent
memory. Available evidence corrects the state, elapsed-time dynamics
propagate it, and a mixture head predicts observations. The dashed loop
propagates the predicted belief when no new observation arrives.
MASS has three observation modalities and seven internal state blocks
($H=256$, $Z=64$); MIMIC has two modalities and two blocks
($H=64$, $Z=32$). These internal partitions are not additional modalities.
The locked MIMIC forecast uses this same history-aware belief.}
\label{fig:network-pipeline}
\end{figure}

\paragraph{Encoding, filtering, and prediction.}
\Cref{fig:network-pipeline} is the module order in the reported checkpoints.
Clinical values, BALF metabolomics, and the imaging vector each pass through
a modality-specific numerical encoder: a masked value, its Boolean mask, and a
learned modality embedding are concatenated and mapped by one linear layer
to width $H$. An all-missing modality writes no evidence.
MASS CT64 is a frozen Qwen2.5-VL latent prepared before training, so the VLM
is not among the $5{,}338{,}172$ learned parameters. MIMIC uses CXR PCA64
the same way. A temporal imaging encoder then maintains an imaging
state of width $H$: four learned queries and four attention heads pool the
current imaging vector, and a GRU cell updates that state from the pooled
token and an elapsed-time feature. On the reported vector interface the
attention has a single key.
Linear maps send the resulting tokens to a $Z$-dimensional mean and diagonal
log variance for each compartment.
The multirate, care-conditioned transition advances those state blocks with a
local MLP, Fourier features of elapsed time ($16$ on Clinical19, $8$ on
MIMIC), and SiLU gates on the fixed cross-compartment edges.
After the last observation, rollout repeats this transition on the predicted
state and decodes with the three-expert observation head. No Transformer
block sits in this path. The causal Transformer and GRU comparators are
trained separately.

The MASS-Syn actor is a categorical policy. It reads the flattened
$3\times 8$ latent state and a four-step one-hot code, $28$ inputs in total,
through two hidden layers of width $128$ with $\tanh$ activations, and it
emits seven action logits. That MLP has $21{,}127$ parameters. A behavior
policy of the same shape is fit by imitation and then held fixed. Imagined
MASS-Syn transitions come from an ensemble of three frozen bilinear
compartment models, each with latent width $8$. This ensemble is the known
simulator's learned transition used to score the same authorized-search
interface. The simulator contains no patient records.

\paragraph{Observation encoder and correction.}
For a modality of width $D_m$ and hidden width $H$, the numeric encoder
concatenates masked values, the $D_m$-dimensional mask, and an
$H$-dimensional learned modality embedding. A linear map
$(2D_m+H)\rightarrow H$ produces its token. An all-missing modality cannot
contribute observed evidence. Each compartment has separate
$H\rightarrow Z$ initial-mean and initial-log-variance projections.
The correction gate is an MLP $(Z+H)\rightarrow H\rightarrow Z$ with
SiLU and sigmoid activations. The innovation projection maps $H\rightarrow Z$;
the predicted-observation projection maps $Z\rightarrow H$.
The innovation is the observed token minus this predicted token. Its gated
projection is added to the prior mean; the prior variance is divided by
one plus the accumulated correction gain. Token availability multiplies the
gain, so an all-missing modality produces no correction. During recursive
execution, decoded means and predicted availability (sigmoid threshold 0.5)
supply the next decoder's source interface without posterior correction.

\paragraph{Dynamics and output heads.}
The Clinical19 graph contains 12 directed edges: reciprocal connections
between global severity and each of the four clinical organ groups,
between imaging and respiration, and between metabolism and inflammation.
Local drift, diffusion, elapsed-time features, and gated messages update
the belief. The learned time gain is bounded, with a 24-hour nominal scale
and a 72-hour evidence half-life in the MASS configuration. Initial
cross-edge probability is 0.05 and residual scale is 0.05.
For an output of width $D_m$, the probabilistic head maps
$Z\rightarrow2D_m$ for dynamic mean/scale, $Z\rightarrow3D_m$ for
three-expert gate logits, and $Z\rightarrow D_m$ for availability.
The experts are persistence, a learned population state, and a dynamic
residual. Persistence is masked out when the source value is absent.

\begin{table}[t]
\centering\small
\caption{\textbf{Checkpoint parameter accounting.} Counts are obtained from
the stored tensor shapes; frozen image backbones, PCA, and target-network
copies are excluded. The history/temporal-image row is the imaging
memory above: multi-head attention plus a GRU cell. The MIMIC parent and action adapter have identical
total size but different optimization scopes. Counts include allocated
modules in the reported checkpoints.}
\label{tab:parameter-breakdown}
\begin{tabular}{@{}lrr@{}}
\toprule
Module & MASS Clinical19 & MIMIC \\
\midrule
Numeric encoders & 537,600 & 18,304 \\
History/temporal image encoder & 939,008 & 62,720 \\
State prior and initial projections & 232,064 & 8,448 \\
Dynamics & 2,649,827 & 62,596 \\
Prediction/innovation/correction & 921,984 & 24,960 \\
Observation and auxiliary heads & 57,689 & 15,433 \\
\midrule
Total & 5,338,172 & 192,461 \\
\bottomrule
\end{tabular}
\end{table}

\subsection{Training phases, losses, and stopping rules}
\label{app:training}

Reported development and locked forecasts use one schedule per cohort
(\cref{tab:model-implementation}). MASS uses the Clinical19 base schedule.
MIMIC uses the history-aware parent schedule and is selected by calibration
recursive balanced RMSE across seeds 20260915--20260917; each selected
checkpoint is from epoch 12. That MIMIC checkpoint uses 13 clinical variables
and CXR for both the development comparison and the locked forecast.
Locked MASS forecasts use the calibration-selected Clinical19 runs.

Constrained refinement and the MIMIC action adapter are later optimization
phases on these checkpoints. They keep the cohort observation setting fixed.
Refinement continues from the Clinical19 checkpoint and updates dynamics,
observation heads, and predicted-observation projections (2,823,851 parameters).
It did not satisfy its acceptance criteria, so it does not replace the locked
MASS forecast. The action adapter updates 20,352 action-related parameters on
the same MIMIC parent, leaves the other modules frozen, and retains epoch 7
by calibration clinical macro RMSE.

%% file: appendix/b_method_details.tex
\subsection{Detailed belief updates, fitting, and policy execution}
\label{app:method-details}

The main text distinguishes observation modalities from the inferred patient
state. This section gives the full parameterization and execution details.
Clinical19 subdivides the clinical observation stream into five physiological
groups; these are internal state blocks, not additional observation modalities.
MIMIC uses two state blocks for its clinical and radiograph observations.
The term compartment below denotes an internal state block.

\subsubsection{Problem setup}
\label{app:detail-problem}

To connect observations collected at different times, we represent
the patient through a belief inferred from the available history.
We model the underlying evolution as a partially observed dynamical
system. Let $t$ index observations at calendar times
$\tau_t$, with elapsed time $\Delta t_t=\tau_{t+1}-\tau_t$.
The ordered history is
$\mathcal H_t=((o_j,m_j,\tau_j,a_{j-1}))_{j=1}^{t}$, where $o_j$ contains
clinical measurements and modality-specific representations, and
$m_j$ identifies the observed entries. The preceding care context
$a_{j-1}$ is included when enabled by the experimental configuration;
future actions are not part of the initializing history.
The model approximates the unobserved patient state $Z_t$
with the learned belief
\begin{equation}
    B_t=q_\theta(Z_t\mid\mathcal H_t).
    \label{eq:belief}
\end{equation}
Missing entries supply no measured value. Their availability remains
explicit rather than being conflated with an observed numerical zero.

The world consists of a belief-transition kernel
$\mathcal T_\theta(B'\mid B,\Delta t,a)$ and an observation model
$\mathcal O_\theta(o,m\mid B)$. The latter predicts observation values
and availability; the former propagates the patient belief, optionally
conditioned on supplied care. Starting from
$\widehat B_t=B_t$, an $H$-step open-loop forecast is defined recursively
by
\begin{align}
    \widehat B_{t+h+1}
    &\sim \mathcal T_\theta\!\left(
    \cdot\mid\widehat B_{t+h},\Delta t_{t+h},a_{t+h}\right),
    \nonumber\\
    (\widehat o_{t+h+1},\widehat m_{t+h+1})
    &\sim \mathcal O_\theta\!\left(
    \cdot\mid\widehat B_{t+h+1}\right),
    \qquad h=0,\ldots,H-1.
    \label{eq:world-rollout}
\end{align}
The kernel notation also accommodates deterministic updates of belief
parameters. No factual intermediate observation is assimilated during
this rollout. Evidence-recency metadata carried alongside the belief
is left implicit in the operator notation.

We distinguish learning these predictive dynamics from optimizing
interventions within them. Conditioning on a supplied action does not
by itself establish its interventional effect. Policy learning is
therefore restricted to actions supported by a separate, fixed
authorization procedure, defined in
\cref{sec:planner}.

\subsubsection{Asynchronous multimodal belief learning}
\label{app:detail-state-estimation}

Asynchronous observations must update the patient state while
retaining information from earlier measurements. We separate
propagation between observations from correction by newly available
evidence, allowing each observation group to refine its corresponding
state block. The Clinical19 instantiation represents this belief with seven
Gaussian state blocks,
\begin{equation}
    B_t=\{B_t^c\}_{c=1}^{7},\qquad
    B_t^c=\mathcal N\!\left(
    \mu_t^c,\operatorname{diag}(\exp\ell_t^c)\right),
    \label{eq:compartment-belief}
\end{equation}
corresponding to respiratory, renal, inflammatory, hemodynamic,
global-severity, lung-imaging, and lung-metabolic states. The first
five organize the 19 clinical variables; the last two represent
CT64 and BALF64. Cohort-specific observation interfaces and the use
of history-enabled filtering are specified in
\cref{sec:experiments,app:architecture}.

Before assimilating an observation at $\tau_t$, the preceding belief
is propagated over $\Delta t_{t-1}$ to obtain a prior
$\widetilde B_t$. Available evidence then corrects this prior
(Figure~\ref{fig:method}(a)). Let $v$ index the observation groups
assigned to state blocks by $c(v)$. Here clinical measurements are
indexed through their five predefined feature groups, rather than
being assigned as a single modality to one state block. Modality
encoders process masked values and masks,
$(m_t^v\odot o_t^v,m_t^v)$, to supply observation tokens $e_t^v$.
The innovation and correction are
\begin{align}
    \nu_t^v
    &=e_t^v-D_v(\widetilde\mu_t^{c(v)}),
    \nonumber\\
    K_t^v
    &=b_t^v\,
    \sigma\!\left(G_v[
    \widetilde\mu_t^{c(v)},\nu_t^v]\right),
    \nonumber\\
    \mu_t^c
    &=\widetilde\mu_t^c+
    \sum_{v:c(v)=c}K_t^v\odot P_v(\nu_t^v),
    \label{eq:belief-correction}
\end{align}
where $b_t^v$ indicates the presence of observed evidence in group
$v$, and $D_v$ and $P_v$ map between belief and token spaces.
The learned gate modulates an observation-driven correction rather
than replacing the propagated state. In particular, $b_t^v=0$
removes that group's measurement correction without preventing
state evolution. The same correction gains reduce the prior variance
elementwise by a factor of one plus the accumulated gain, while
learned nonnegative diffusion increases variance during propagation.
Alongside this distribution, each state block carries evidence
reliability and observation age. Observed evidence resets reliability
to one and age to zero; between observations, reliability decays
exponentially with a prescribed half-life and age increases with time.
Compartments without initial evidence start with zero reliability.
Thus, evidence age measures time since the last observation and is
distinct from the elapsed time between consecutive transitions.
The complete updates are specified in \cref{app:architecture}.

\subsubsection{Shared-state dynamics and world-model fitting}
\label{app:detail-world-dynamics}

Recursive simulation carries each state block forward after observations stop. Each block combines a local transition with sparse cross-block messages:
\begin{align}
    d_t^c
    &=F_{\mathrm{loc}}^c(\mu_t^c,T(\Delta t_t))
      +\sum_{j\rightarrow c}
      g_{j\rightarrow c,t}M_{j\rightarrow c}(\mu_t^j),
    \nonumber\\
    \eta_c(\Delta t_t)
    &=1-\exp\!\left[
    -\operatorname{softplus}(\omega_c)\Delta t_t/24
    \right],
    \nonumber\\
    \widetilde\mu_{t+1}^c
    &=\mu_t^c+\eta_c(\Delta t_t)d_t^c.
    \label{eq:natural-dynamics}
\end{align}
Here $T$ supplies Fourier time features, $\Delta t_t$ is expressed
in hours, and $g_{j\rightarrow c,t}$ is a learned message gate.
In the evidence-gated configuration, it is scaled by source
reliability and a target-receptivity factor that increases as target
evidence becomes stale. A separate posterior-reliability option
scales messages using the source and target mean log variances.
These quantities describe evidence freshness and within-model belief
uncertainty, respectively; the ensemble uncertainty used for planning
is defined below. The gain satisfies $\eta_c(0)=0$.
The sparse graph couples physiological state blocks through global
severity and links imaging to respiration and metabolism to
inflammation. These connections are the message graph of the transition.

To compare supplied care conditions from a common patient state,
we branch this same transition through a state-dependent action
residual. The residual is initialized to zero and augments the shared transition when care is supplied:
\begin{equation}
    \widetilde\mu_{t+1}^c
    =\mu_t^c+\eta_c(\Delta t_t)
    \left[
    d_t^c+
    g_c(\mu_t^c,E_a(a_t),\Delta t_t)R_c(E_a(a_t))
    \right].
    \label{eq:care-context}
\end{equation}
The encoder $E_a$ represents care, while $g_c$ determines its
state-dependent contribution. Omitting this residual defines
state-only prediction. Observation heads decode clinical variables and
modality-specific representations, with separate heads for future
availability. Imaging outputs are latent representations rather
than synthesized CT or CXR pixels. After initialization, predicted
beliefs supply subsequent transitions as in
\cref{eq:world-rollout}; decoded predictions are not treated as
new factual measurements.

\paragraph{World-model fitting.}
Training grounds the shared belief in observed measurements and
supervises its evolution through recursive predictions.
T1 supervises source-state reconstruction, while T2 supervises
recursive future predictions and availability
(the lower panel of Figure~\ref{fig:method}). Writing the corresponding terms as
$\mathcal L_{\mathrm{rec}}$, $\mathcal L_{\mathrm{dyn}}$, and
$\mathcal L_{\mathrm{avail}}$, world fitting can be summarized as
\begin{equation}
    \mathcal L_{\mathrm{world}}
    =\lambda_{\mathrm{rec}}\mathcal L_{\mathrm{rec}}
    +\lambda_{\mathrm{dyn}}\mathcal L_{\mathrm{dyn}}
    +\lambda_{\mathrm{avail}}\mathcal L_{\mathrm{avail}}
    +\mathcal L_{\mathrm{aux}}.
    \label{eq:world-fitting}
\end{equation}
The active terms and weights follow the stage-specific schedules
in \cref{app:training}. Reconstruction hides available source entries
and minimizes their Gaussian negative log-likelihood under the
observation heads. Recursive supervision applies the corresponding
masked likelihood to future values, with binary cross-entropy for
their availability. Value losses are evaluated only where target
measurements are observed; future observations provide supervision,
not intermediate rollout inputs.
On MIMIC, three reconstruction epochs can precede nine rollout epochs within the same twelve-epoch budget.
The history comparison crosses that history-aware input with
an auxiliary reconstruction objective.
$\mathcal L_{\mathrm{aux}}$ collects posterior-alignment, direct-path
consistency, and grounding terms where applicable. The MIMIC rollout
objective also supervises direct forecasts and aligns recursive
beliefs with stopped-gradient direct predictions using smooth-$L_1$
losses. MASS-Syn fits each ensemble member by one-step latent-state
mean squared error and selects its checkpoint on calibration data.
After fitting and calibration,
$\theta$ is frozen. Prognostic probes evaluate the information in
predicted states and do not train the world dynamics.

The formulation above covers the framework's components; \cref{sec:experiments} specifies their use. MIMIC development comparisons and the locked forecast use one history-aware belief over 13 clinical variables and CXR. MASS-Syn instantiates action-conditioned
dynamics and policy improvement on simulator latent states.

\subsubsection{Evidence-constrained model-based reinforcement learning}
\label{app:detail-planner}

Action-conditioned futures become useful for planning when evidence
supports comparing the consequences of the candidate actions.
Cap-MBRL makes this requirement explicit by determining action
admissibility before optimizing behavior. Authorization restricts
the policy's action support; uncertainty-penalized policy learning
and risk-budgeted selection govern improvement within that support
(Figure~\ref{fig:method}(b)).

\paragraph{Evidence-based action authorization.}
The authorization audit tests both whether recorded actions inform
prediction and whether their effects support interventional comparison.
T3 denotes this prespecified audit, separate from world-model fitting.
Let $S_t$ denote recorded source-state features used for
auditing, distinct from the learned belief. Predictive evidence
compares registered state-only and action-conditioned predictors,
$f(S_t,\Delta t)$ and $f_A(S_t,\Delta t,a_t)$, on
patient-disjoint data. An audit cell specifies the action comparison
covered by a certificate; MASS-Syn uses one cell per intervention
channel. An audited cell must satisfy
\begin{equation}
    \Delta_A=\mathcal L(f)-\mathcal L(f_A)\geq\epsilon_A,
    \qquad
    p_{A,\mathrm{adj}}\leq\alpha_A,
    \label{eq:predictive-gate}
\end{equation}
and pass the prespecified action-shuffling and support checks.
Here $p_{A,\mathrm{adj}}$ summarizes patient-clustered evidence for
the required tests. MASS-Syn evaluates ridge and extra-trees
predictors, with Bonferroni adjustment for family selection and
Holm correction across action cells. Together these conditions
define $G_A$. A separate indicator $G_I$
records whether the identification evidence required for
interventional comparison is available and passes the prescribed
checks. Predictive improvement alone never sets $G_I=1$.
In MASS-Syn, known transition means provide the reference action
effect relative to no intervention. Identification checks bound
effect-prediction error and binary action-versus-reference decision
regret, and require sufficient oracle gain and patient support.
Patient-clustered tests with Holm correction determine which cells
pass these checks on calibration data.
The resulting authorized set is
\begin{equation}
    \mathcal A_{\mathrm{auth}}
    =\{a:G_A(B,a)=1,\ G_I(B,a)=1\}.
    \label{eq:authorized-set}
\end{equation}
The gate notation refers to fixed audit decisions. In MASS-Syn,
these decisions are indexed by action channel and held constant
throughout a rollout, so no mapping from imagined beliefs to
source-state bins is required. The no-intervention reference $a_0$
is always retained, giving
$\mathcal A_{\mathrm{auth}}^{\mathrm{Syn}}
=\{a_0\}\cup\{a_j:G_A(j)G_I(j)=1\}$.
Channels without a passing certificate remain excluded at every
imagined step. State-dependent extrapolation is assessed through
the risk signals below, rather than through a new state-specific
certificate. The audit protocol specifies support conditions and
identification checks (\cref{app:execution}).

When the set is nonempty, the actor softmax is normalized only over
authorized actions at each imagined step. An empty set produces
abstention rather than an unauthorized control rollout. In this
study, policy learning and selection are confined to known-mechanism
MASS-Syn; real-cohort action conditioning remains predictive.

\paragraph{Policy learning from imagined trajectories.}
Within the authorized action set, policy learning evaluates choices
through their cumulative predicted consequences. Comparing continuations
from the same initial state provides the learning signal while
holding the patient's starting condition fixed.
On MASS-Syn, the actor $\pi_\phi$ generates $K$ continuations from
each initial belief $B_i$ in a frozen three-member world ensemble.
The members share a transition architecture but use separate random
initializations and training orders. Their mean prediction supplies
the next latent state. At each step, $u$ is the variance across
member predictions, averaged over latent coordinates and normalized
by its calibration-set 90th percentile. It measures ensemble
disagreement and does not include the Gaussian belief variance.
Let $\mathbb P_{\theta,\phi}$ denote the resulting trajectory
distribution. For continuation $k$, its uncertainty-penalized return
and group-normalized advantage are
\begin{align}
    S_{ik}
    &=\sum_{h=1}^{H}\gamma^{h-1}
    (r_{ikh}-\kappa u_{ikh}),
    \nonumber\\
    \widehat A_{ik}
    &=\frac{S_{ik}-\overline S_i}
    {\max(\operatorname{Std}_k(S_{ik}),\epsilon_{\mathrm{adv}})},
    \label{eq:group-advantage}
\end{align}
where $u_{ikh}$ is model uncertainty and $\overline S_i$ averages
returns across continuations from the same initial belief.
Rewards are derived from the MASS-Syn latent-severity objective;
the reward convention and uncertainty statistic are specified in
\cref{app:planner-implementation}.
A behavior policy $\mu_b$ is first fitted by cross-entropy to the
logged seven-way action and then frozen. These actions comprise no
intervention and one unit intervention in each of six channels;
simultaneous multi-channel actions are outside this policy interface.
The actor is initialized from the behavior policy. The implemented
sampled objective is
\begin{align}
    \mathcal L_{\mathrm{Cap}}
    ={}&-\mathbb E_{i,k}\!\left[
    \operatorname{sg}(\widehat A_{ik})
    \sum_{h=1}^{H}\gamma^{h-1}
    \log\pi_\phi(a_{ikh}\mid B_{ikh})
    \right]
    \nonumber\\
    &+\lambda_{\mathrm{KL}}
    \widehat D_{\mathrm{KL}}(\pi_\phi,\mu_b)
    +\lambda_\rho
    \widehat{\mathbb E}_{i,k}
    \left[(\bar\rho_{ik}-\bar b_{ik})_{+}\right],
    \label{eq:cgrpi}
\end{align}
where $\operatorname{sg}$ denotes the stop-gradient operator on the advantage.
The risk statistic combines ensemble disagreement, behavior
surprisal, recursive--direct discrepancy, and capability-violation
diagnostics. Behavior surprisal is the negative log-probability of
the chosen action under $\mu_b$, divided by $\log 7$.
Path discrepancy is the mean squared difference between recursive
states and a separately fitted direct predictor conditioned on the
action history, normalized on calibration data. Each nonnegative
signal $x$ is mapped to $x/(1+x)$ before weighted averaging;
$\bar\rho$ and $\bar b$ are discount-normalized trajectory averages.
The capability-violation signal is zero for authorized samples.

Imagined rewards enter the actor update as a score-function gradient: the stopped advantage weights the log-probability of the categorical sample. The KL term and the positive risk-budget gap are part of the same sampled objective.
A nonnegative multiplier tracks mean budget violation and scales
$\lambda_\rho$. The same budget is enforced again when the learned policy
is used for decision-time selection.
The per-step budget is
\begin{equation}
    b_h=\operatorname{clip}
    (b_0-\alpha\rho_h^{\mathrm{ref}},b_{\min},b_0).
    \label{eq:risk-budget}
\end{equation}
During training, $\rho_h^{\mathrm{ref}}$ evaluates a no-intervention
step from the current imagined state, retaining the preceding action
history. At decision time, the reference is the candidate branch
whose first action is $a_0$ and whose continuation follows the actor.
The base decision-time budget is calibrated to a prespecified
acceptance rate and then frozen. This reference is separate from
the fallback trajectory used to assess improvement.

Only $\phi$ is optimized: world parameters, audit decisions, the
behavior model, and the fallback remain fixed. The exhaustive
minimizer used to evaluate selection regret is not a policy-training
target.

\paragraph{Risk-constrained decision-time selection.}
At decision time, the learned policy supplies continuations for
comparing immediate actions, while the risk budget and fallback
comparison determine whether to adopt a proposal.
MASS-Syn selection enumerates the seven possible first actions and
restricts selection to authorized candidates; training samples
continuations instead. After fixing the first action,
each candidate branch uses the trained policy's highest-probability
authorized action at every subsequent imagined state.
Thus, policy learning determines the continuation used to evaluate
each first action. A separate fallback branch follows its own fixed
policy throughout the remaining horizon. In the reported experiment,
this is the capability-mask policy, initialized from $\mu_b$ and
trained within the same authorized action set before being frozen.

Let $\widehat C(B,a)$ and $\widehat\rho(B,a)$ denote a candidate's
discounted cost and discount-normalized risk. The cost sums predicted
latent severity and a weighted risk penalty over the remaining horizon.
Let $\widehat b(B)$ be the shared state-dependent budget, and
$\widehat C_{\mathrm{fb}}(B)$ the cost of the fallback's own continuation.
Its first action $a_{\mathrm{fb}}$ must itself be authorized.
If no authorized fallback exists, selection abstains.
Otherwise, for a frozen improvement margin $\delta\geq0$,
define
\begin{align}
    \mathcal F(B)=\bigl\{a\in\mathcal A_{\mathrm{auth}}:\;&
    \widehat\rho(B,a)\leq\widehat b(B),
    \nonumber\\
    &\widehat C(B,a)<
    \widehat C_{\mathrm{fb}}(B)-\delta\bigr\}.
    \label{eq:feasible-improvements}
\end{align}
The executed action is
\begin{equation}
    a_{\mathrm{exec}}(B)=
    \begin{cases}
        \bot,
        &a_{\mathrm{fb}}\notin\mathcal A_{\mathrm{auth}},\\
        \displaystyle\arg\min_{a\in\mathcal F(B)}\widehat C(B,a),
        &a_{\mathrm{fb}}\in\mathcal A_{\mathrm{auth}},\;
        \mathcal F(B)\neq\varnothing,\\
        a_{\mathrm{fb}},&\text{otherwise},
    \end{cases}
    \label{eq:constrained-selection}
\end{equation}
where $\bot$ denotes abstention; an unavailable fallback is treated
as outside the authorized set. In MASS-Syn, retaining $a_0$ ensures
that the masked fallback has an admissible action.
The budget screens proposals that would replace the fallback;
fallback execution itself is not conditional on satisfying that budget.
The selected first action is executed only in MASS-Syn, and the
comparison repeats at the next state. The continuation rule and
cost definitions are part of the fixed decision-time protocol
(\cref{app:planner-implementation}).

%% file: tables/tab_action_ontology.tex
\begin{table}[t]
\centering
\small
\caption{Frozen MASS action ontology. Twenty privacy-safe macro classes are
collapsed by a within-group maximum into ten mechanism blocks.}
\label{tab:action-ontology}
\begin{tabular}{p{0.27\linewidth}p{0.66\linewidth}}
\toprule
Mechanism block & Member macro classes \\
\midrule
Broad-spectrum beta-lactam & Antipseudomonal beta-lactam/inhibitor,
carbapenem, cephalosporin \\
Resistant Gram-positive & Gram-positive antibiotic \\
Other antibacterial & Fluoroquinolone, tetracycline, aminoglycoside,
polymyxin, other antimicrobial \\
Antifungal & Antifungal \\
Antiviral & Antiviral \\
Immunomodulation & Systemic corticosteroid, other immunomodulator \\
Nutrition/microbiome support & Enteral nutrition, parenteral nutrition,
probiotic \\
Acid suppression & Acid suppression \\
Ventilation support & Invasive mechanical ventilation \\
Extracorporeal support & Continuous renal replacement therapy, ECMO \\
\bottomrule
\end{tabular}
\end{table}

%% file: appendix/c_additional_results.tex
\FloatBarrier
\section{Complete Quantitative Results}
\label{app:results}
This section supplements the main experiments with full forecasting
comparisons, source-reconstruction and history controls, and action-evidence
diagnostics. Each comparison identifies its evaluation population: locked
clinical tests, development component studies, or known-mechanism simulation.
These populations address complementary questions and are not pooled.

\subsection{Evaluation metrics and uncertainty}
\label{app:experiment-metrics}
Macro NMAE averages feature-wise errors normalized by training IQR.
Consecutive-change error is
\begin{equation}
\operatorname{NMAE}_{\Delta}
=\frac{|(\hat y_h-\hat y_{h-1})-(y_h-y_{h-1})|}
{\operatorname{IQR}_{\rm train}},\qquad \hat y_0=y_0.
\label{eq:step-change-nmae}
\end{equation}
For complete trajectories with identical feature weights and normalization,
this definition implies that H4 NMAE cannot exceed four times the mean
H1--H4 change error. Some Direct-VLM entries in
\cref{tab:t2-complete} do not satisfy this relation at the aggregate level.
The reported aggregates do not establish whether endpoint and change errors
use identical eligible cells and weights; their comparability therefore
remains unresolved. Low change error in these rows should not be interpreted
as evidence of accurate temporal dynamics without a matched-support check.
H4 direction agreement measures sign agreement, not change magnitude.
EHR-S@15 averages feature-level success at relative error at most 15\%,
with denominator $\max(|y|,0.1\operatorname{IQR}_{\rm train})$.
It is a retrospective compatibility score, not a clinical acceptability threshold.
Whole-patient bootstrap uses 2,000 resamples for forecasting
and action evaluation, and 5,000 for seed-averaged MASS trends and paired
future-state prognosis comparisons. Component-specific resampling protocols
are stated with their tables. 

The unauthorized-action rate is the fraction of executed actions outside the permitted set.
A zero rate means the hard mask was respected. It does not measure whether
an authorized comparison was identified. Selection
regret compares decisions against the best action in the permitted set. Full definitions
are in \cref{app:planner-implementation}.

\subsection{Complete locked-test forecasting comparison}
\label{app:locked-forecast}
\input{tables/tab_t2_complete}

\paragraph{Benchmark conventions.}
\label{app:benchmark-conventions}
CLARITY is adapted separately for each cohort. On MASS, that comparator
is fitted to clinical variables and CT64, without BALF, and uses calibration
selection. SFT is fitted separately on the clinical forecast interface:
Clinical19 for MASS and the 13-variable plus CXR interface for MIMIC.
MASS persistence carries the last observed interface forward with elapsed
time retained; train median uses the train-split median vector. Both adapted
models use the cohort's shared outcome probe. \method{} forecast and prognosis
remain on the reported checkpoints: Clinical19 with CT and BALF for MASS,
and 13 clinical variables with CXR for MIMIC (\cref{app:vlm-prognosis}).

\FloatBarrier

\input{appendix/c_experiment_support}

\subsection{T1 source completion}
\label{app:t1-state}

T1 here is masked source completion on the numerical interface used by the
forecasting baselines. The frozen \method{} checkpoint is evaluated on exactly
the same 271 MIMIC requests, 718 hidden cells, train-only transforms, and
scorer as the Direct VLMs. Requested current clinical values are zeroed and
masked before encoding; the action-disabled model then decodes the masked
belief after a $10^{-3}$-h transition. This is a source-completion comparison.
T1 scores masked source completion on this checkpoint, not the three-seed mean used for locked recursive forecasts. The history-versus-fusion
comparison is the development experiment in \cref{sec:state-utility}.

\input{tables/tab_t1_locked_main}
\input{tables/tab_t1_vlm}
\FloatBarrier

\paragraph{MASS source-reconstruction validation.}
A separate MASS Task~1 validation evaluates 105 held-out examples (210 masked
source values) under the Clinical19 reconstruction curriculum
(\cref{tab:t1-main}). It obtains masked NMAE $0.667$,
masked Gaussian NLL $3.228$, 95\% interval coverage $90.0\%$, and availability Brier score $0.278$. Transferring this value
into the MASS panel above would mix distinct masks, examples, and scoring
protocols.

\input{appendix/c_mechanism_evidence}

\subsection{MIMIC predictive action conditioning}
\label{app:mimic-pred-action}

The MIMIC action-conditioned \method{} is a supervised world model whose input
includes logged fluid and vasopressor doses. \Cref{tab:t3-action} reports
factual-versus-zero and shuffled-action checks together with the authorization
gate. Relative RMSE
reductions versus the frozen state-only parent are $0.20\%$ on all validation
windows, $0.61\%$ on action-exposed windows, and $0.79\%$ at H4.

\input{tables/tab_t3_main}
\input{tables/tab_t3_action}
\FloatBarrier

\begin{figure*}[!htbp]
  \centering
  \includegraphics[width=\textwidth]{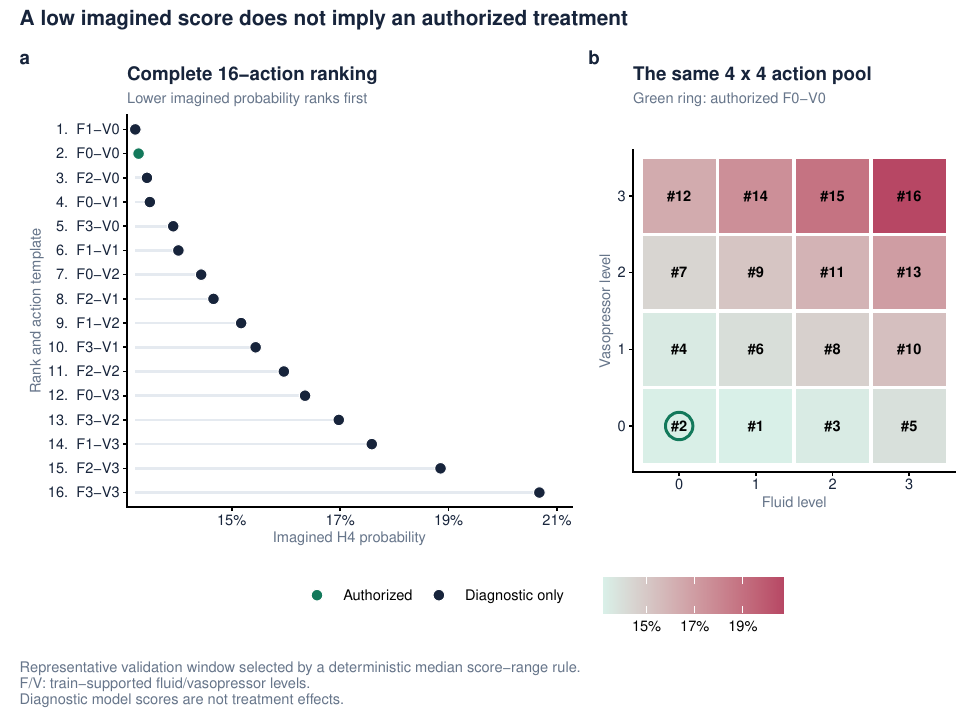}
  \caption{\textbf{A complete single-window ranking of all 16 MIMIC action
  templates.} (a) The frozen action-conditioned model ranks every train-derived
  fluid--vasopressor template by imagined H4 mortality probability for one
  representative validation window. The case is selected deterministically as
  the median score-range window among those whose imagined optimum is not the
  no-op fallback template. (b) The same ranks in the $4\times4$ action
  grid. F1--V0 has the lowest model score, and the displayed branch retains F0--V0.}
  \label{fig:mimic-action-ranking-case}
\end{figure*}
\FloatBarrier

\subsection{Authorization-gated selection on known-mechanism worlds}
\label{app:cgrpi}

The policy comparison scores selection after the identification gate opens on MASS-Syn.
The confirmatory experiment compares unrestricted and penalty-based world-model
RL, an authorized fallback, an authorized soft-penalty policy, and risk-budgeted
policy improvement (\cref{tab:cgrpi,fig:action-planning-summary}). The aggregate across five
independent environments met the prespecified confirmatory criteria;
individual environments are replicates, not separate confirmatory studies.

\input{tables/tab_cgrpi}

\paragraph{Interpreting the policy controls.}
The ungated controls test whether uncertainty, support, or path-consistency
penalties can replace explicit authorization. Their lower unrestricted
regret is not evidence of better selection within the authorized set.
The gated comparison holds that set fixed. Cap-MBRL reduces mean selection regret from
0.0910 to 0.0180.
Across the five environments, selection regret improves in four and is
unchanged in one. 

\FloatBarrier

%% file: tables/tab_t2_complete.tex
\begin{table}[H]
\centering
\caption{\textbf{Locked-test longitudinal simulation.}
H1--H4 report macro NMAE; $\Delta$-NMAE scores consecutive changes
(patient-cluster $95\%$ CIs); H4 dir.\ scores direction agreement.
Bold and underline mark the best and second-best task-trained result in each column. Ties share marks.
}
\label{tab:t2-complete}
\scriptsize
\setlength{\tabcolsep}{2.4pt}
\renewcommand{\arraystretch}{0.82}

{\footnotesize\textbf{MIMIC-IV-CXR pneumonia locked test (300 patients)}}\\[2pt]
\resizebox{\textwidth}{!}{%
\begin{tabular}{@{}lccccccc@{}}
\toprule
\rowcolor{tblhead}
Method & H1$\downarrow$ & H2$\downarrow$ & H3$\downarrow$ & H4$\downarrow$ & H4 EHR-S@15$\uparrow$ & $\Delta$-NMAE$\downarrow$ & H4 dir.$\uparrow$ \\
\midrule
\rowcolor{tblhead!45}
\multicolumn{8}{@{}l}{\textbf{Controls}} \\
Persistence & 0.6740 & 0.7491 & 0.7692 & 0.7752 & 49.4\% & 0.443 [0.425,0.461] & 56.8\% \\
Train median & 0.7188 & 0.7009 & 0.6839 & 0.7090 & 48.8\% & 0.471 [0.454,0.489] & 59.6\% \\
\addlinespace[0pt]
\rowcolor{tblhead!45}
\multicolumn{8}{@{}l}{\textbf{Temporal models}} \\
GRU & 0.5068 & \textbf{0.5483} & \underline{0.5818} & \underline{0.5934} & 52.1\% & 0.412 [0.396,0.428] & 60.4\% \\
Causal Transformer & \textbf{0.5005} & 0.5527 & 0.5877 & 0.5954 & \underline{52.5\%} & \underline{0.411 [0.395,0.428]} & 60.1\% \\
GRU-D & 0.5238 & 0.6815 & 0.8418 & 0.9600 & 38.3\% & 0.444 [0.429,0.461] & 58.1\% \\
\addlinespace[0pt]
\rowcolor{tblhead!45}
\multicolumn{8}{@{}l}{\textbf{General-purpose Direct VLMs}} \\
Qwen2.5-VL-3B$^{\dagger}$ & 0.6740 & 0.7491 & 0.7692 & 0.7752 & 49.4\% & 0.443 [0.425,0.461] & 56.8\% \\
Qwen2.5-VL-7B & 2.0176 & 2.8403 & 2.1843 & 1.8233 & 44.7\% & 0.682 [0.572,0.795] & 56.7\% \\
Qwen2.5-VL-72B & 0.9069 & 2.2291 & 0.8848 & 3.0042 & 44.0\% & 0.774 [0.446,1.410] & 57.0\% \\
Qwen3-VL-8B-Instruct & 4.5230 & 5.5157 & 6.1501 & 6.2845 & 46.0\% & 0.467 [0.449,0.486] & 57.0\% \\
Qwen3-VL-30B-A3B & 33.2483 & 35.0294 & 40.3677 & 25.2325 & 45.2\% & 0.581 [0.469,0.754] & 56.1\% \\
InternVL3-8B$^{\dagger}$ & 0.6740 & 0.7491 & 0.7692 & 0.7752 & 49.4\% & 0.443 [0.425,0.461] & 56.8\% \\
LLaVA-OV-1.5-8B & 2.4461 & 1.0740 & 3.4511 & 1.2832 & 38.3\% & 0.644 [0.564,0.742] & 55.5\% \\
\addlinespace[0pt]
\rowcolor{tblhead!45}
\multicolumn{8}{@{}l}{\textbf{Medical Direct VLMs}} \\
Lingshu-7B & 1.0245 & 1.1025 & 1.1310 & 1.1103 & 46.9\% & 0.458 [0.430,0.500] & 56.8\% \\
Hulu-Med-7B & 24.1641 & 11.6647 & 14.1120 & 9.8788 & 47.0\% & 0.897 [0.564,1.500] & 57.7\% \\
Hulu-Med-32B & 54.5972 & 34.4492 & 45.6867 & 22.0711 & 47.8\% & 0.588 [0.447,0.845] & 57.6\% \\
\addlinespace[0pt]
\rowcolor{tblhead!45}
\multicolumn{8}{@{}l}{\textbf{Task-adapted VLM and medical world model}} \\
Qwen2.5-VL-7B SFT & 0.6484 & 1.6620 & 0.7182 & 0.7221 & 49.5\% & 0.875 [0.589,1.415] & 56.4\% \\
CLARITY (adapted)$^{\P}$ & 0.5110 & 0.5746 & 0.6129 & 0.6328 & 50.8\% & 0.531 [0.513,0.550] & \underline{61.4\%} \\
\addlinespace[0pt]
\rowcolor{tblhead!45}
\multicolumn{8}{@{}l}{\textbf{Ours}} \\
\rowcolor{tblours}
\method{} & \underline{0.5034} & \underline{0.5489} & \textbf{0.5722} & \textbf{0.5780} & \textbf{54.9\%} & \textbf{0.407 [0.392,0.423]} & \textbf{61.7\%} \\
\bottomrule
\end{tabular}}

\vspace{2pt}
{\footnotesize\textbf{MASS Clinical19 locked test (40 patients; three training seeds)}}\\[2pt]
\resizebox{\textwidth}{!}{%
\begin{tabular}{@{}lccccccc@{}}
\toprule
\rowcolor{tblhead}
Method & H1$\downarrow$ & H2$\downarrow$ & H3$\downarrow$ & H4$\downarrow$ & H4 EHR-S@15$\uparrow$ & $\Delta$-NMAE$\downarrow$ & H4 dir.$\uparrow$ \\
\midrule
\rowcolor{tblhead!45}
\multicolumn{8}{@{}l}{\textbf{Controls}} \\
Persistence & 0.5726 & 0.6727 & 0.5951 & 0.5563 & 38.5\% & 0.523 [0.449,0.610] & 50.0\% \\
Train median & 0.5444 & 0.6134 & 0.5672 & 0.5563 & 38.5\% & 0.509 [0.439,0.578] & 50.0\% \\
\addlinespace[0pt]
\rowcolor{tblhead!45}
\multicolumn{8}{@{}l}{\textbf{Temporal models}} \\
GRU & 0.5438 & 0.6648 & 0.5984 & 0.6908 & 39.7\% & 0.507 [0.445,0.577] & 52.2\% \\
Causal Transformer & 0.5543 & 0.7039 & 0.6080 & 0.6995 & 35.0\% & \underline{0.503 [0.440,0.572]} & 52.2\% \\
GRU-D & 0.5237 & \underline{0.6397} & \underline{0.5718} & \underline{0.5442} & 42.2\% & \underline{0.503 [0.436,0.575]} & \textbf{56.7\%} \\
\addlinespace[0pt]
\rowcolor{tblhead!45}
\multicolumn{8}{@{}l}{\textbf{General-purpose Direct VLMs}} \\
Qwen2.5-VL-3B & 2.3613 & 2.8903 & 6.8647 & 16.4983 & 3.8\% & 0.527 [0.454,0.616] & 63.3\% \\
Qwen2.5-VL-7B & 0.7887 & 1.2337 & 0.8895 & 1.0127 & 30.4\% & 0.561 [0.479,0.651] & 60.0\% \\
Qwen2.5-VL-72B & 4.9553 & 0.9430 & 0.8230 & 0.9236 & 36.4\% & 0.532 [0.453,0.624] & 50.0\% \\
Qwen3-VL-8B-Instruct & 0.8699 & 1.1391 & 0.8965 & 1.0233 & 45.5\% & 0.525 [0.450,0.610] & 60.0\% \\
Qwen3-VL-30B-A3B & 2.9213 & 3.1071 & 6.2477 & 1.3304 & 29.1\% & 0.526 [0.453,0.611] & 60.0\% \\
InternVL3-8B$^{\dagger}$ & 0.5726 & 0.6727 & 0.5951 & 0.5563 & 38.5\% & 0.523 [0.449,0.610] & 50.0\% \\
LLaVA-OV-1.5-8B & 6.3862 & 6.7173 & 17.9637 & 18.4743 & 0.0\% & 0.558 [0.475,0.661] & 66.7\% \\
\addlinespace[0pt]
\rowcolor{tblhead!45}
\multicolumn{8}{@{}l}{\textbf{Medical Direct VLMs}} \\
Lingshu-7B & 4.8341 & 2.8278 & 9.5505 & 10.0853 & 11.2\% & 2.390 [0.915,4.076] & 66.7\% \\
Hulu-Med-7B & 6.3582 & 6.6926 & 17.8894 & 18.4743 & 0.0\% & 0.522 [0.449,0.610] & 66.7\% \\
Hulu-Med-32B & 0.9901 & 1.2992 & 0.8678 & 1.0012 & 44.3\% & 0.534 [0.463,0.618] & 56.7\% \\
\addlinespace[0pt]
\rowcolor{tblhead!45}
\multicolumn{8}{@{}l}{\textbf{Task-adapted VLM and medical world model}} \\
Qwen2.5-VL-7B SFT & 4.8415 & 1.3606 & 2.2658 & 1.5103 & 26.2\% & 2.519 [1.625,3.660] & \underline{53.3\%} \\
CLARITY (adapted)$^{\P}$ & \textbf{0.5217} & 0.8358 & 0.6554 & 0.6302 & \textbf{49.9\%} & 0.601 [0.542,0.664] & \underline{53.3\%} \\
\addlinespace[0pt]
\rowcolor{tblhead!45}
\multicolumn{8}{@{}l}{\textbf{Ours}} \\
\rowcolor{tblours}
\method{} & \underline{0.5218} & \textbf{0.5997} & \textbf{0.5658} & \textbf{0.5418} & \underline{46.8\%} & \textbf{0.498 [0.431,0.569]} & 51.1\% \\
\bottomrule
\end{tabular}}

\vspace{2pt}
\parbox{0.98\textwidth}{\footnotesize
$^{\P}$CLARITY is independently adapted and fitted per cohort.
$^{\dagger}$This Direct VLM row repeats the persistence control of its panel on every cell.
MASS and MIMIC task-trained models and \method{} are three-seed means.
EHR-S is a retrospective compatibility score, not a clinical acceptability
threshold.}
\end{table}

%% file: appendix/c_experiment_support.tex
\subsection{Future-state prognosis and extended rollout displays}
\label{app:extended-rollout-evidence}

One frozen outcome probe per cohort reads future-state summaries. The probe
is shared across methods within a cohort and is not shared between cohorts.
MIMIC locked test has 300 pneumonia patients and 32 in-hospital deaths, scored at H4.
MASS uses a separate 99-patient frozen-validation panel with 28 deaths by
day 28, scored at the available horizon: H2, H3, and H4 for 2, 46, and 51
patients. Fallback handling is in \cref{app:benchmark-conventions}.
Relative to the current-state control, predicted states raised AUROC from
$0.7106$ to $0.7135$ on MIMIC and from $0.6856$ to $0.7186$ on MASS.
AUPRC and Brier score also improved.
The complete prognosis comparison is retained in the main text
(\cref{tab:t4-full}), alongside the MASS clinical and representation
forecast display (\cref{fig:forecast-comparison}).

\Cref{fig:asynchronous-evidence}a shows observed CXR references; panel (b)
shows lower \method{} representation error than CLARITY and persistence on
that window.
Panels (c--e) show selected clinical trajectories after observations stop.
The CT64 and BALF64 development displays in \cref{fig:forecast-comparison}
are separate from locked-test clinical scores. H1 MSE was $0.836/0.836$,
compared with $0.868/0.889$ for GRU. These checks concern representation forecasts.

The case figure is retained in the main text (\cref{fig:asynchronous-evidence}).
Its CXR panel compares \method{}, CLARITY, persistence, and the train mean
in the same standardized representation space, not pixel space.
The clinical panels include observed values, \method{}, the causal Transformer,
GRU, and persistence. The six-hour labels denote event windows rather than
exact image acquisition times (\cref{fig:same-patient-io}).
Case selection and interpretation are detailed in \cref{app:integrated-trajectories}.
\FloatBarrier

%% file: tables/tab_t1_locked_main.tex
\begin{table}[t]
\centering
\caption{\textbf{Masked source reconstruction under two evaluation protocols.}
The MIMIC column is the frozen locked test: 271 requests, 718 masked cells,
the same train-only IQR and scorer as the Direct VLMs in
\cref{tab:t1-vlm}. The MASS column is the Clinical19 development validation: 105 examples and 210 masked values. It is not the 40-patient locked forecast, so the columns are not a paired comparison. Lower NMAE, RMSE, NLL, and Brier score
are better. Coverage closer to 95\% is better.}
\label{tab:t1-main}
\footnotesize
\setlength{\tabcolsep}{4pt}
\renewcommand{\arraystretch}{1.05}
\begin{tabular}{@{}lcc@{}}
\toprule
 & MIMIC locked test & MASS Task 1 validation \\
\midrule
Requests / masked values & 271 / 718 & 105 / 210 \\

Masked NMAE & 0.783 cell; 0.834 macro & 0.667 \\
Std.\ RMSE / Gaussian NLL & 1.143 & 3.228 \\
EHR-S@10/15/25 & 34.6 / 42.8 / 58.8 & -- \\
95\% coverage / availability Brier & -- & 90.0\% / 0.278 \\
\bottomrule
\end{tabular}
\end{table}

%% file: tables/tab_t1_vlm.tex
\begin{table}[H]
\centering
\caption{\textbf{T1 source completion on the numerical interface.}
MIMIC-IV-CXR uses the frozen locked-test cohort (271 requests; 718 scored
cells), a 25\% stateless observed-cell mask, and train-only median/IQR
normalization. The \method{} row uses that same mask, cohort, and scorer.
MASS uses the Clinical19 locked-test protocol (20 requests; 37 scored
cells) and has no matched \method{} row.  EHR-S is feature-macro in both panels.}
\label{tab:t1-vlm}
\scriptsize
\setlength{\tabcolsep}{3.2pt}
\renewcommand{\arraystretch}{1.04}

\textbf{MIMIC-IV-CXR locked test}\\[-1pt]
\resizebox{0.94\textwidth}{!}{%
\begin{tabular}{@{}lrrrr@{}}
\toprule
Method & Cell NMAE$\downarrow$ & Macro NMAE$\downarrow$
& Std. RMSE$\downarrow$ & EHR-S@10/15/25 (\%$\uparrow$) \\
\midrule
\method{} (frozen) & \textbf{0.783} & \textbf{0.834} & \textbf{1.143} & 34.6 / 42.8 / 58.8 \\
Qwen2.5-VL-3B & 7.834 & 7.029 & 69.054 & 18.4 / 22.4 / 30.5 \\
Qwen2.5-VL-7B & 7.764 & 6.419 & 17.717 & 12.4 / 16.1 / 21.2 \\
Qwen2.5-VL-72B & 15.103 & 14.914 & 123.536 & 20.1 / 24.5 / 32.5 \\
Qwen3-VL-8B-Instruct & 21.294 & 22.602 & 163.906 & 29.2 / 35.8 / 48.3 \\
Qwen3-VL-30B-A3B & 26.089 & 25.263 & 146.563 & 7.2 / 8.5 / 11.8 \\
InternVL3-8B & 1.338 & 1.304 & 5.524 & 35.5 / 43.5 / 55.9 \\
LLaVA-OV-1.5-8B & 13.884 & 13.429 & 101.546 & 14.9 / 17.6 / 23.0 \\
Lingshu-7B & 9.233 & 8.559 & 80.423 & 19.7 / 24.1 / 33.2 \\
Hulu-Med-7B & 6.452 & 6.333 & 70.518 & 31.4 / 38.8 / 51.1 \\
Hulu-Med-32B & 114.072 & 118.842 & 362.066 & 29.6 / 36.5 / 49.8 \\
\bottomrule
\end{tabular}%
}

\vspace{6pt}
\textbf{MASS Clinical19 locked test}\\[-1pt]
\resizebox{0.94\textwidth}{!}{%
\begin{tabular}{@{}lrrrr@{}}
\toprule
Method & Cell NMAE$\downarrow$ & Macro NMAE$\downarrow$
& Std. RMSE$\downarrow$ & EHR-S@10/15/25 (\%$\uparrow$) \\
\midrule
Qwen2.5-VL-3B & 4.327 & 4.895 & 13.644 & 16.1 / 16.1 / 18.0 \\
Qwen2.5-VL-7B & 17.607 & 21.532 & 100.550 & 24.5 / 40.5 / 51.8 \\
Qwen2.5-VL-72B & 1.556 & 1.494 & 1.874 & 20.0 / 23.3 / 33.3 \\
Qwen3-VL-8B-Instruct & 4.041 & 4.573 & 13.587 & 16.1 / 16.1 / 16.1 \\
Qwen3-VL-30B-A3B & 6.309 & 7.350 & 19.122 & 10.0 / 15.6 / 15.6 \\
InternVL3-8B & 0.572 & 0.592 & 0.757 & 43.0 / 56.2 / 69.4 \\
LLaVA-OV-1.5-8B & 1.867 & 1.899 & 2.207 & 16.7 / 19.4 / 22.2 \\
Lingshu-7B & 1.491 & 1.528 & 1.859 & 26.7 / 33.7 / 35.5 \\
Hulu-Med-7B & 0.573 & 0.593 & 0.758 & 43.0 / 56.2 / 69.4 \\
Hulu-Med-32B & 1.451 & 1.466 & 1.798 & 22.2 / 22.2 / 31.1 \\
\bottomrule
\end{tabular}%
}

\vspace{2pt}
\parbox{0.98\textwidth}{\footnotesize
Standardized RMSE uses train-IQR normalization. The MIMIC \method{} row uses the same frozen history-aware checkpoint: requested current clinical values are hidden before encoding and decoded after
an effectively zero-time ($10^{-3}$ h) transition.  Bold marks the lowest
error in the MIMIC panel. These are source-completion diagnostics, not
recursive forecasting or control comparisons.}
\end{table}

%% file: appendix/c_mechanism_evidence.tex
\subsection{Component ablations and missing-evidence sensitivity}
\label{app:mechanism-evidence}
The component studies test three distinct capabilities: retaining observation
history, predicting clinical values and modality representations from a shared state, and selecting actions under
fixed authorization. The following map links each capability to its controls
and states the interpretation supported by those controls.
\begin{center}
\small\renewcommand{\arraystretch}{1.15}
\begin{tabular}{@{}p{.23\textwidth}p{.36\textwidth}p{.35\textwidth}@{}}
\toprule
Design & Corresponding evidence & Boundary \\
\midrule
Asynchronous observation updates & MIMIC development history comparison (\cref{tab:history-retention}) & On MIMIC, history-aware filtering improves prognosis versus latest-value fusion, with and without auxiliary reconstruction. \\
Multimodal evolving state & CXR/CT/BALF representation prediction; clinical rollouts; frozen-probe prognosis & The shared state supports clinical forecasts and CT/CXR/BALF representation forecasts under the same evaluation. \\
Treatment-conditioned transitions & Logged-versus-zero/shuffled conditioning and same-source candidate branches & Logged care changes the predicted trajectory. \\
Frozen-world planning & Authorized-set regret and risk-budget comparison on MASS-Syn & Authorized search lowers selection regret in MASS-Syn. \\
\bottomrule
\end{tabular}
\end{center}

The main text summarizes history retention in \cref{tab:history-retention}. Reconstruction-curriculum and missing-evidence controls are reported below.

\subsubsection{History-enabled MIMIC prognosis follow-up}
\label{app:mimic-history-prog}
The development comparison uses one frozen H4-eligible window per patient on the MIMIC development set (3,758 patients; 524 in-hospital deaths), and three training seeds. History-aware filtering updates the state from
ordered observations. Latest-value fusion merges each channel to its most
recent value. The comparison is reported in \cref{tab:history-retention}:
without reconstruction, filtering improves AUROC by $0.0213$
($95\%$ CI $[0.0132,0.0295]$). With reconstruction, the AUROC, AUPRC, and
Brier changes relative to latest-value fusion are $+0.0381$, $+0.0778$,
and $-0.0110$. Both comparisons use this MIMIC history-aware belief.

\subsubsection{MIMIC curriculum and missing-evidence controls}
The development controls use the same MIMIC history-aware model as the
locked forecast. Reconstruction replaces the first three of twelve training
epochs rather than adding compute. Inference-only dropout uses probabilities
25\%, 50\%, and 75\% for each observed clinical value and, independently,
for each whole CXR channel. Patient windows are paired across arms, with
three training seeds. No locked-test data are used.

\begin{table}[!htbp]
\centering\small
\caption{\textbf{Matched component ablations on MIMIC development.}
Top: clinical RMSE, mean $\pm$ sample SD across three training seeds.
Bottom: paired patient-level RMSE reduction; positive favors the named
component. Each patient's window/horizon MSE is averaged before taking
the square root, then averaged across seeds. These contrasts therefore
have a different aggregation from the top-panel scores. Intervals use
5,000 patient-bootstrap resamples over 4,036 patients, conditional on
the three seeds.}
\label{tab:matched-ablation}
\label{tab:recon-curriculum}
\begin{tabular}{@{}llr@{}}
\toprule
Time input & Training / input arm & Clinical RMSE $\downarrow$\\
\midrule
Elapsed & Rollout only & $0.7688\pm0.0022$\\
Elapsed & Reconstruction + rollout & $0.7646\pm0.0021$\\
\midrule
\multicolumn{2}{l}{Paired component contrast} & Gain [95\% CI]\\
\midrule
\multicolumn{2}{l}{Reconstruction curriculum} & $0.00371\ [0.00196,0.00535]$\\
\bottomrule
\end{tabular}
\end{table}

\begin{table}[!htbp]
\centering\small
\caption{\textbf{Inference with missing source evidence.} Clinical RMSE,
mean $\pm$ sample SD over the same three seeds. Source dropout masks each
observed clinical value and drops each whole CXR channel at the stated
probability, without retraining. Perturbations are paired across arms.
These rows mask observed clinical values and drop whole CXR channels together.
They are development robustness diagnostics.}
\label{tab:missing-evidence}
\setlength{\tabcolsep}{5pt}
\begin{tabular}{@{}lcc@{}}
\toprule
Source perturbation & Elapsed-time rollout & With reconstruction\\
\midrule
None & $0.7688\pm0.0022$ & $0.7646\pm0.0021$\\
25\% source dropout & $0.8264\pm0.0010$ & $0.8239\pm0.0015$\\
50\% source dropout & $0.8891\pm0.0007$ & $0.8884\pm0.0036$\\
75\% source dropout & $0.9504\pm0.0015$ & $0.9500\pm0.0055$\\
\bottomrule
\end{tabular}
\end{table}
\FloatBarrier

\subsection{Task-adapted VLM prognosis evaluation}
\label{app:vlm-prognosis}
Qwen2.5-VL-7B SFT is trained and scored on the clinical forecast interface: Clinical19 targets for MASS, and the 13-variable plus CXR interface for MIMIC.

%% file: tables/tab_t3_main.tex
\begin{table*}[!htbp]
\centering
\caption{\textbf{From action-conditioned prediction to authorized policy
improvement.} The capability ladder separates an action representation,
predictive action information ($G_A$), causal identification ($G_I$), and
policy optimization. MASS-Syn tests Cap-MBRL after known-mechanism
identification. MIMIC candidates are means of train-transition action vectors
within train-fitted dose cells; no non-train row contributes to a prototype.}
\label{tab:t3-main}
\scriptsize
\setlength{\tabcolsep}{3.0pt}
\renewcommand{\arraystretch}{1.12}

{\footnotesize\textbf{Capability and evidence ladder}}\\[2pt]
\begin{tabular}{@{}>{\raggedright\arraybackslash}p{0.075\textwidth}>{\raggedright\arraybackslash}p{0.145\textwidth}>{\raggedright\arraybackslash}p{0.15\textwidth}>{\raggedright\arraybackslash}p{0.11\textwidth}>{\raggedright\arraybackslash}p{0.135\textwidth}>{\raggedright\arraybackslash}p{0.295\textwidth}@{}}
\toprule
\rowcolor{tblhead}
World & Action representation & Action evidence & Identification $G_I$
& Authorized treatment set & Planner / policy result and claim \\
\midrule
Real MASS & 10 mechanism blocks; 8 prespecified cells
& $0/8$ cells pass the full gate
& Closed: observational
& $\varnothing$
& Eight prespecified cells; planning is scored in MASS-Syn \\
MIMIC-IV-CXR & $4\times4$ fluid--vasopressor cells; 16 train-only prototypes
& $+0.61\%$ RMSE gain on exposed windows; $+0.79\%$ at H4; logged beats zero/shuffles
& Closed: no causal design
& $\varnothing$; no-action fallback $a_0$
& Logged care changes the forecast; no nonzero action is authorized \\
MASS-Syn & Known-mechanism action space
& $3/3$ signals open; $3/3$ matched nulls close
& Ground-truth audit
& Nonempty where supported
& Cap-MBRL regret $0.0180$ vs. fallback $0.0910$; unauthorized action rate $0$ \\
\bottomrule
\end{tabular}

\vspace{2pt}
\parbox{0.97\textwidth}{\footnotesize
The MIMIC candidate grid, dose thresholds, and replay protocol are in
\cref{tab:t3-action}. Since $G_I=0$, no non-zero candidate is authorized.
The MASS-Syn value $0$ is the rate of executed actions outside the permitted set.
Bootstrap intervals and the MIMIC planner audit are in
\cref{tab:t3-action}.}
\end{table*}

%% file: tables/tab_t3_action.tex
\begin{table}[!htbp]
\centering
\scriptsize
\setlength{\tabcolsep}{3.6pt}
\caption{\textbf{MIMIC action interface: prediction is enabled, control fails
closed.} Fluid and vasopressor doses are discretized on
the train split into 16 joint templates ($4\times4$; thresholds $5.83$/$12.36$
ml/kg and $0.065$/$0.148$ mcg/kg/min NEE). Template 0 is no-action ($75.8\%$ of
train transitions). Each candidate prototype is the within-cell mean of only
those train-transition action vectors; calibration, validation, and test rows
are excluded from prototype construction. Identification contrasts reuse the
frozen action-conditioned \method{} on the same
2048-window / 1380-patient validation subsample as Cap-MBRL
(without accessing the test set). Relative RMSE reduction compares logged doses
with the named contrast on action-exposed windows ($n=750$);
the state-only parent is the frozen natural-history model, and the
all-window parent reduction is $0.20\%$. Positive values mean logged doses are more predictive.
The ranking searches the 16 templates by imagined death probability.}
\label{tab:t3-action}
{\footnotesize\textbf{A. Frozen candidate-action pool}}\\[2pt]
\begin{tabular}{@{}lcccc@{}}
\toprule
Fluid $\backslash$ vasopressor & None & Low & Medium & High \\
\midrule
None   & $a_0$  & $a_1$  & $a_2$  & $a_3$ \\
Low    & $a_4$  & $a_5$  & $a_6$  & $a_7$ \\
Medium & $a_8$  & $a_9$  & $a_{10}$ & $a_{11}$ \\
High   & $a_{12}$ & $a_{13}$ & $a_{14}$ & $a_{15}$ \\
\bottomrule
\end{tabular}

\vspace{4pt}
{\footnotesize\textbf{B. Predictive action-information contrasts (action-exposed)}}\\[2pt]
\begin{tabular}{@{}lcc@{}}
\toprule
Contrast vs logged doses & All-H RMSE red.\ $\uparrow$ & H4 RMSE red.\ $\uparrow$ \\
\midrule
State-only parent & $0.61\%$ $[0.45,0.80]$ & $0.79\%$ $[0.59,1.03]$ \\
Zero action & $0.82\%$ $[0.24,1.68]$ & $1.26\%$ $[0.39,2.37]$ \\
Global shuffle & $0.47\%$ $[0.18,0.79]$ & $0.73\%$ $[0.24,1.28]$ \\
Support-matched shuffle & $0.48\%$ $[0.16,0.84]$ & $0.66\%$ $[0.21,1.18]$ \\
\bottomrule
\end{tabular}

\vspace{4pt}
{\footnotesize\textbf{C. Fail-closed policy interface on the same windows}}\\[2pt]
\begin{tabular}{@{}lcccc@{}}
\toprule
Method & Treat.\ rate & Unauth.\ & Imagined death $P$ & $\Delta$ vs current \\
\midrule
Cap-MBRL (hard gate) & \textbf{0.000} & \textbf{0.000} & 0.136 & $-0.013$ \\
Logged factual sequence (audit) & 0.366 & 0.366 & 0.139 & $-0.010$ \\
Unrestricted diagnostic planner & 0.740 & 0.740 & 0.125 & $-0.024$ \\
\bottomrule
\end{tabular}
\end{table}

%% file: tables/tab_cgrpi.tex
\begin{table*}[!htbp]
\centering
\footnotesize
\setlength{\tabcolsep}{2.4pt}
\caption{\textbf{Authorization-gated world-model RL on known-mechanism MASS-Syn.}
Means over five independent simulator-parameter and split seeds in the
confirmatory experiment; the aggregate met the prespecified criteria.
Lower is better. Selection regret is policy error within the authorized set. Rows
with nonzero unauthorized action rates can choose actions outside the permitted set, so their
selection regret is not comparable. The authorized soft-penalty policy combines
the mask with uncertainty, support, and path-consistency terms, without
Cap-MBRL's budgeted candidate selection. Among rows with
zero unauthorized action rate, bold and underline mark the best and second-best
selection regret. Rates are fractions, not percentages.}
\label{tab:cgrpi}
\begin{tabular}{@{}lcc@{}}
\toprule
Method & Selection regret & Unauth.\ rate \\
\midrule
\multicolumn{3}{@{}l}{\textit{Diagnostic controls without hard authorization}}\\
Unrestricted WM-RL & -- & 0.3219 \\
Uncertainty-aware & -- & 0.3226 \\
Support-aware & -- & 0.3371 \\
Path-consistency weighting & -- & 0.3158 \\
\midrule
\multicolumn{3}{@{}l}{\textit{Policy comparison within the same authorized action set}}\\
Authorized + soft penalties & 0.3636 & \textbf{0.0000} \\
Authorized-set fallback & \underline{0.0910} & \textbf{0.0000} \\
\rowcolor{tblours} \textbf{Cap-MBRL} & \textbf{0.0180} & \textbf{0.0000} \\
\midrule
No action & -- & 0.0000 \\
\bottomrule
\end{tabular}
\end{table*}

%% file: appendix/c_action_visuals.tex
\clearpage
\subsection{Action-conditioned branches and authorization evidence}
\label{app:action-visuals}
The main-text case uses the first identity-hash-ordered validation window
with all three displayed source variables observed (212 eligible windows
among 2,048). Selection does not use future measurements or model outputs.
Fluid templates pool recorded crystalloid boluses (normal saline and lactated
Ringer's); vasopressors use norepinephrine-equivalent rates, not a
drug-specific intervention. For Actions 1--4, the displayed fluid/rate pairs
are $(0,0)$, $(0,0.252)$, $(18.710,0)$, and $(20.904,0.292)$, respectively,
in mL/kg and $\mu$g/kg/min. These are inverse transforms of mean
training-set templates in model space, not mean administered doses.
The combined template therefore differs slightly in dose from the two
single-channel templates.
The same template is applied in every six-hour rollout interval.

\begin{figure}[H]
  \centering
  \includegraphics[width=\textwidth]{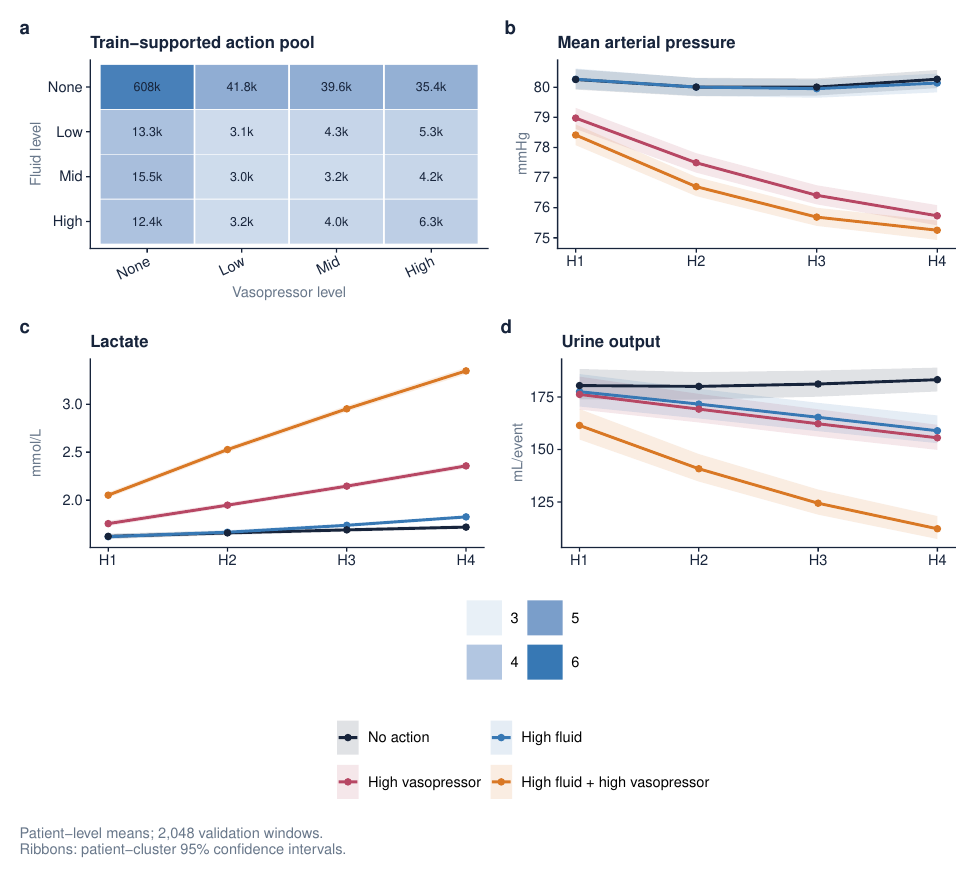}
  \caption{\textbf{What an action is and how it changes a frozen simulation.}
  (a) The MIMIC action pool is a $4\times4$ train-supported grid over fluid and
  vasopressor levels; annotations give train-transition counts. The four
  corners define the diagnostic candidates replayed in panels (b--d).
  Repeating these four train-derived templates over H1--H4 separates predicted mean
  arterial pressure, lactate, and urine-output trajectories. Lines are
  patient-level means over 2,048 validation windows and bands are
  patient-cluster $95\%$ CIs. This action sensitivity is predictive input to
  the gate.}
  \label{fig:mimic-action-trajectories}
\end{figure}

\FloatBarrier

%% file: appendix/b_raw_cases.tex
\clearpage
\section{Case Atlas: Inputs, Future States, and Failure Modes}
\label{app:raw-cases}
\subsection{Matched source and next-state records}
\label{app:matched-records}

The case atlas connects observed clinical and imaging evidence to model
inputs, recursive predictions, and their limitations. It first presents
matched records, then a complete same-patient forecast, followed by
additional clinical, imaging, and metabolomic examples.
The images are observed scans, not reconstructions or model generations. The
numbers are canonical physical-unit values before train-fitted normalization;
``missing'' remains missing through an explicit mask. Exact drug names and
identifiers are mapped to privacy-safe action families before modeling.

\begin{figure}[H]
  \centering
  \includegraphics[width=\textwidth]{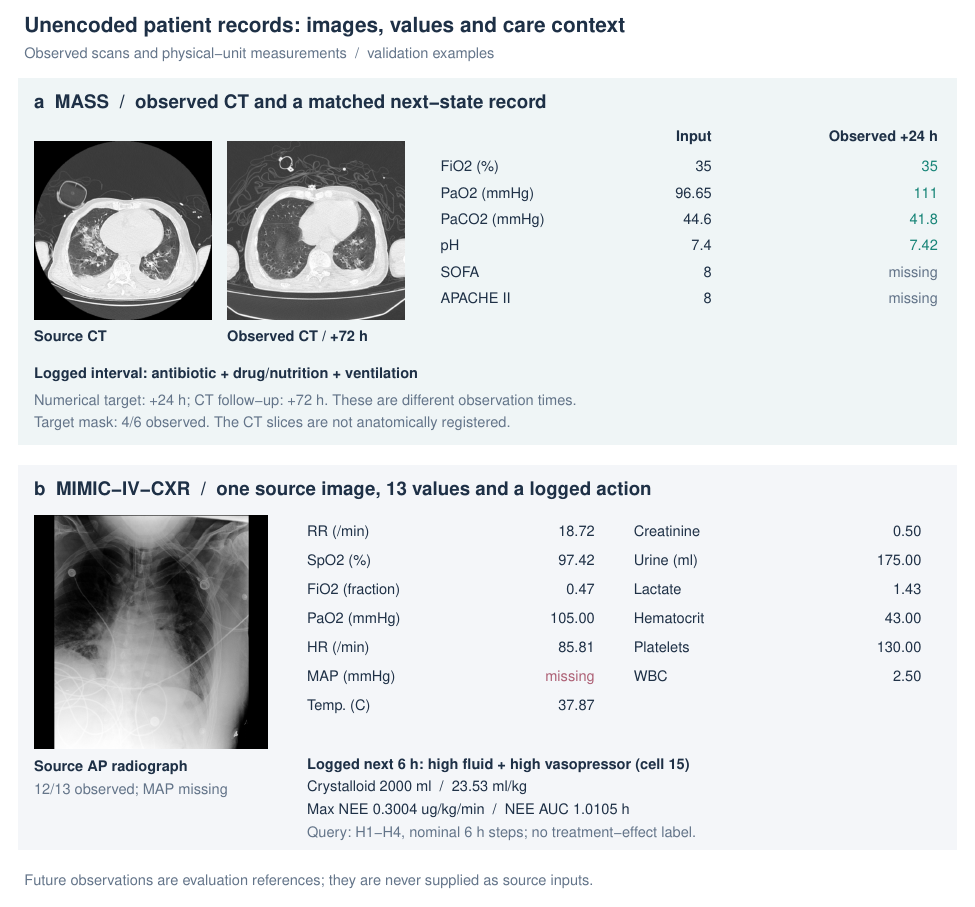}
  \caption{\textbf{Integrated pre-encoder case records.}
  (a) A MASS validation CT pair is shown beside its matched Clinical19
  source and observed next-state record. Numerical references are at +24 h,
  whereas the observed CT follow-up is at +72 h. Missing target cells remain
  missing; neither images nor target values are generated. The fixed
  lung-window slices need not depict identical anatomy and do not support
  pixelwise comparison. The full Clinical19 field list is in
  \cref{tab:raw-case-inputs}.
  (b) A MIMIC validation radiograph, all 13 source clinical values, the
  missing MAP mask, and its logged fluid/vasopressor context are grouped
  together. Images retain their grayscale intensities; fitting and padding
  preserve aspect ratio. Future observations are evaluation references only.}
  \label{fig:raw-case-inputs}
\end{figure}

\clearpage
\subsection{Clinical and action records before encoding}

The examples below report the measurements supplied to the model.
They preserve physical units, missing-value masks, event timing, and action
semantics; they do not contain PCA coordinates, normalized values, or latent
states. MASS exposes all 19 source clinical variables, whereas MIMIC-IV-CXR
exposes 13 variables and retains a false mask for the missing MAP cell.

\begin{table}[H]
\centering
\caption{\textbf{Full pre-encoder clinical records.} MASS shows the
Clinical19 extraction (19/19 observed), the same interface as the
action evaluation record in \cref{fig:raw-case-inputs}; MIMIC shows the 13-field source
record in that figure (12/13 observed). Missing MAP remains masked.
Units are shown where documented for these records.}
\label{tab:raw-case-inputs}
\small
\renewcommand{\arraystretch}{1.08}
\setlength{\tabcolsep}{7pt}
\begin{tabular}{@{}lrlr@{}}
\toprule
\multicolumn{2}{c}{MASS: Clinical19} &
\multicolumn{2}{c}{MIMIC-IV-CXR: 13 clinical} \\
\cmidrule(lr){1-2}\cmidrule(lr){3-4}
Variable & Source value & Variable & Source value \\
\midrule
FiO$_2$ (fraction) & 0.35 & FiO$_2$ (fraction) & 0.467 \\
PaO$_2$ (mmHg) & 91.3 & PaO$_2$ (mmHg) & 105 \\
PaCO$_2$ (mmHg) & 44.6 & SpO$_2$ (\%) & 97.42 \\
pH & 7.40 & Respiratory rate (/min) & 18.72 \\
Respiratory rate (/min) & 12 & Heart rate (/min) & 85.81 \\
Lactate (mmol/L) & 1.50 & Lactate & 1.43 \\
MAP (mmHg) & 70 & MAP (mmHg) & missing \\
Heart rate (/min) & 165 & Temperature ($^\circ$C) & 37.87 \\
Temperature ($^\circ$C) & 35.3 & Creatinine & 0.50 \\
Platelets & 28 & Platelets & 130 \\
WBC & 9.37 & WBC & 2.50 \\
Creatinine & 88 & Urine output (ml) & 175 \\
Bilirubin & 16.4 & Hematocrit & 43 \\
SOFA & 8 & & \\
Renal SOFA & 0 & & \\
APACHE II & 8 & & \\
CRP & 28.7 & & \\
PCT & 0.45 & & \\
BUN & 13.02 & & \\
\midrule
Observed mask & 19/19 & Observed mask & 12/13 \\
\bottomrule
\end{tabular}
\end{table}

\paragraph{Time and action interface.}
The MASS action evaluation example queries the next event after 24 h under logged
antibiotic, drug/nutrition, and ventilation context; the supported alternative
retains drug/nutrition and ventilation only and is a simulation query without an
observed counterfactual target. MIMIC queries H1--H4 at nominal 6 h steps.
Its logged cell 15 contains high fluid and high vasopressor exposure:
2000 ml crystalloid (23.53 ml/kg), maximum NEE
0.3004 $\mu$g/kg/min, and NEE AUC 1.0105 h. Alternative candidates come from
the train-supported $4\times4$ pool; ranking is diagnostic unless the
authorization gate opens. Radiology reports are not model inputs.

\subsection{From raw records to model interfaces}
MASS supplies Clinical19 values and masks, an optional CT, an optional BALF
metabolomic profile, elapsed time, and action context at each asynchronous
event. The full action representation has 20 macro classes with
presence, interval-overlap fraction, source-time activity, and
within-interval initiation channels; the
action labels shown here form the coarser case-display interface. MIMIC supplies
13 clinical values and masks, an optional CXR, elapsed time, and four continuous
fluid/vasopressor quantities mapped to a train-fitted 16-cell candidate pool.
The displayed reports aid human audit only. Action fields in both cohorts record logged care.
\FloatBarrier

\input{appendix/d_worked_outputs}

\clearpage
\subsection{Diverse source, follow-up, and action-context examples}
\label{app:diverse-cases}
\begin{figure}[H]
\centering
\includegraphics[width=\textwidth]{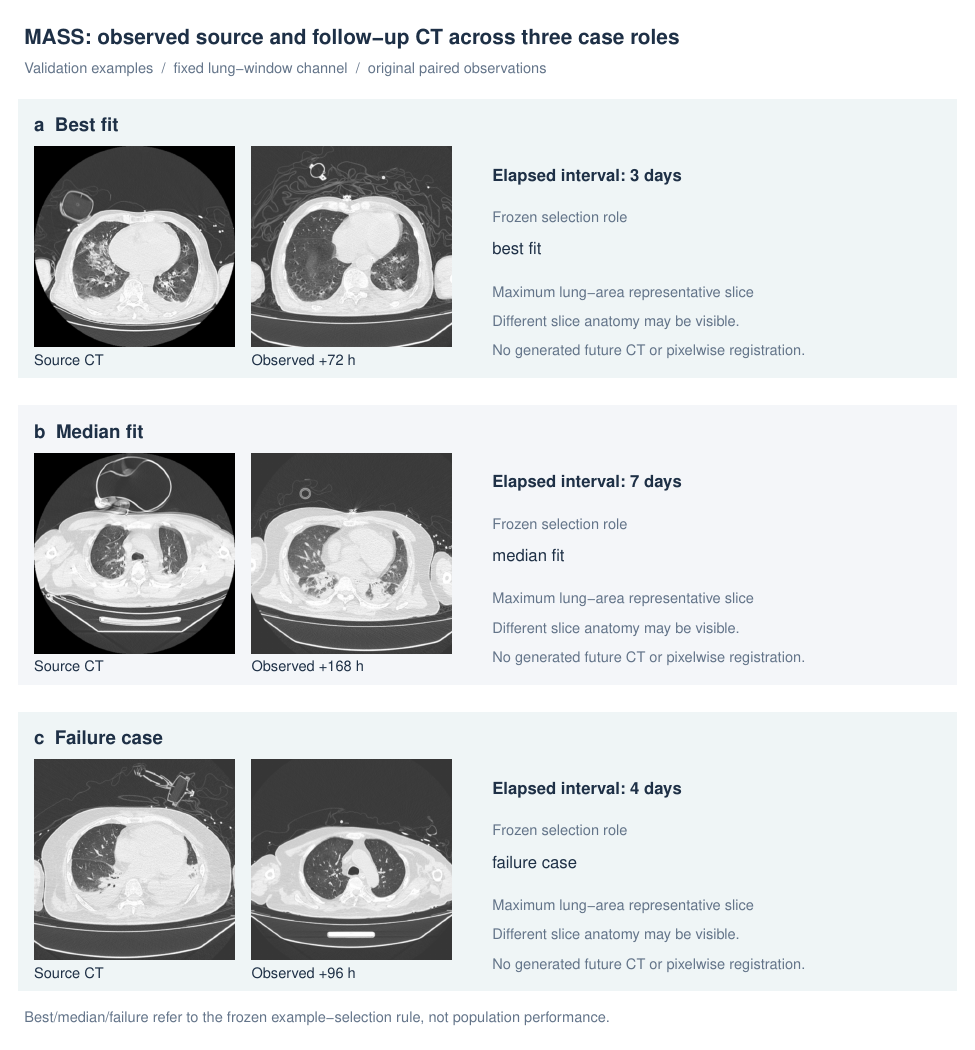}
\caption{\textbf{Three MASS temporal CT examples.} The frozen best-fit,
median-fit, and failure-case roles are all retained. Source and future scans
are observed validation images, separated by 3, 7, and 4 days, respectively.
Only the original lung-window channel is displayed. Representative slices
are not anatomically registered.}
\label{fig:mass-case-atlas}
\end{figure}

\clearpage
\begin{figure}[H]
\centering
\includegraphics[width=\textwidth]{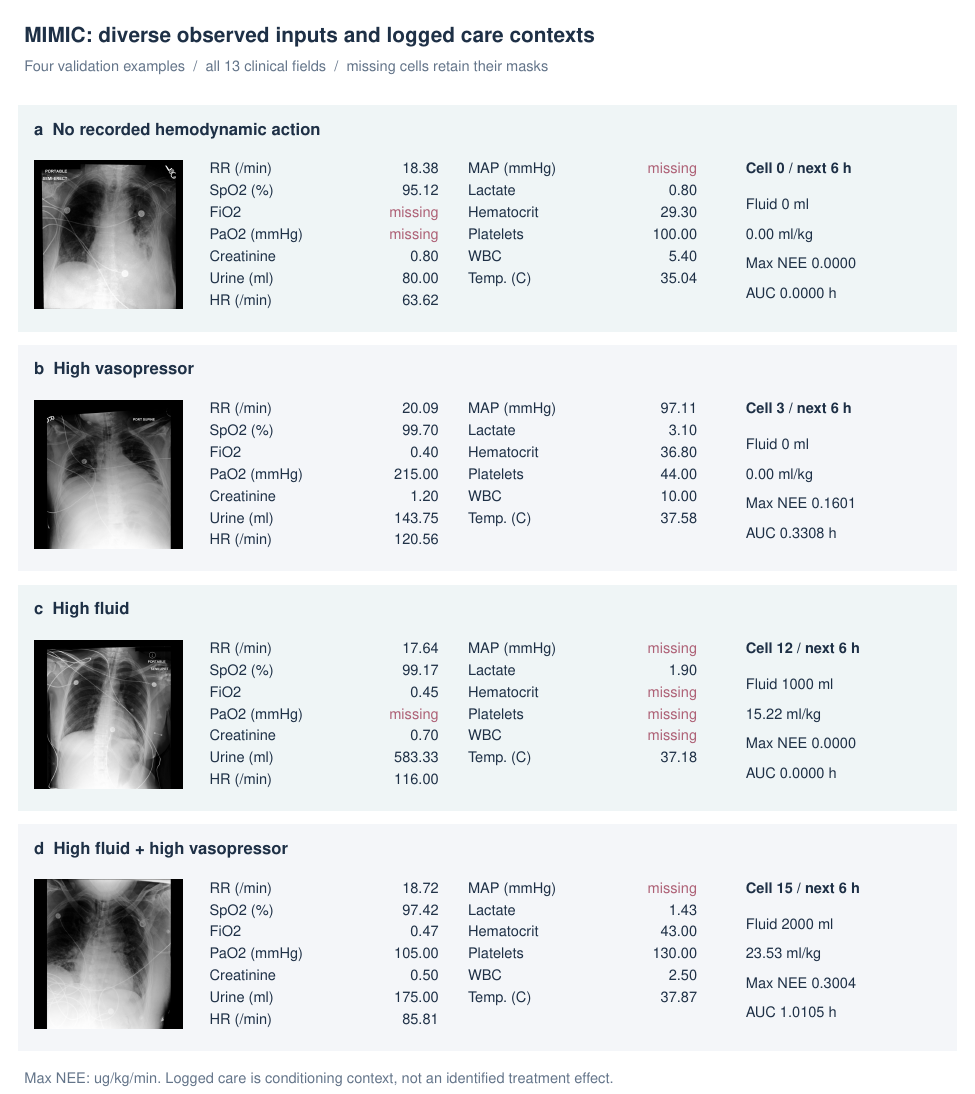}
\caption{\textbf{Four MIMIC input cases with different logged care contexts.}
Each row combines the original source radiograph, all 13 clinical values,
missingness, and the following six-hour action record from the same
validation case. The examples cover no recorded hemodynamic action and
different fluid/vasopressor exposures. Dose summaries are factual context;
these rows contain no observed counterfactual outcomes or treatment
recommendations. They are distinct from the main-text forecasting cases.}
\label{fig:mimic-input-atlas}
\end{figure}
\FloatBarrier

%% file: appendix/d_worked_outputs.tex
\clearpage
\subsection{A complete same-patient multimodal input--output example}
\label{app:worked-outputs}
\label{app:same-patient-io}

This example links the existing complete-CXR validation window to all clinical
inputs and forecasts from the same patient, admission, and source state.
The window was selected for radiograph coverage, not prediction accuracy.
The source contains seven observed clinical variables, six masked variables,
and an observed CXR representation.

\begin{figure}[H]
\centering
\includegraphics[width=\textwidth]{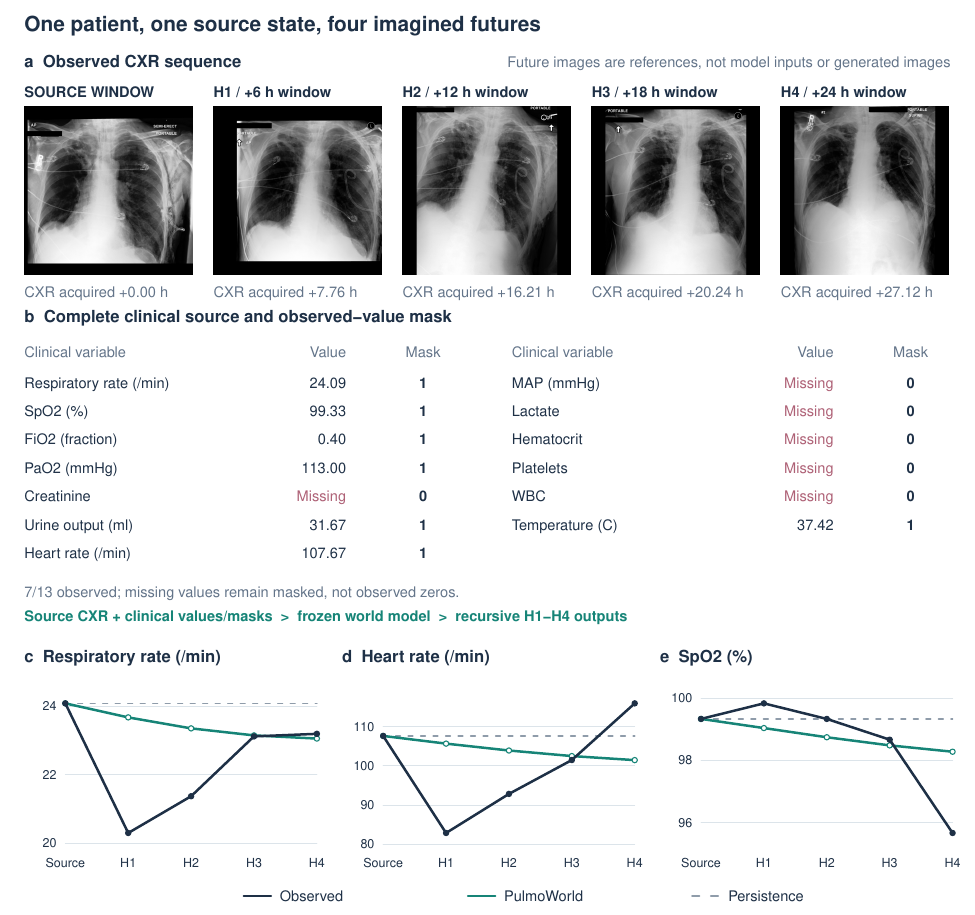}
\caption{\textbf{From one multimodal source record to recursive future outputs.}
(a) Observed AP radiographs from the source and four future windows.
Acquisition offsets are relative to the source radiograph, whereas horizon
labels denote nominal six-hour event-window offsets. One image is displayed
per window; the underlying image counts are 1, 1, 2, 2, and 3.
The model uses the pooled window representation, not only the displayed view.
(b) All 13 clinical input fields and their observed-value masks.
(c--e) Three clinical coordinates decoded from the same four-step rollout,
with observed references and source-value persistence. All other coordinates
are retained in \cref{tab:worked-forecast-output}.
Only the source observations initialize the rollout; future images and
measurements serve as evaluation references. Images retain their grayscale
intensities with aspect-preserving fitting and padding. This single-patient illustration has no calibrated individual prediction intervals.}
\label{fig:same-patient-io}
\end{figure}

\clearpage
\paragraph{Complete numerical outputs and missing-reference handling.}
The frozen natural-history model executes four six-hour transitions without
future observation updates. Its zero model-action vectors are an evaluation
setting, not evidence that the patient received no treatment.
\Cref{tab:worked-forecast-output} reports every clinical decoder mean, including
coordinates whose future reference is unavailable. Missing references are
never replaced with zero for evaluation.

\begin{table}[H]
\centering\footnotesize
\caption{\textbf{Complete clinical outputs for the same multimodal case.}
Each future entry is prediction / observed reference (P / O); ``--'' means
unobserved, with reference mask zero. Source values use physical units before
normalization. Decoder means are shown even when predicted availability is
below 0.5; these means are not treated as available inputs at the next step.
Unrounded values, normalized inputs, masks, availability probabilities, and
observed-cell errors are retained in the accompanying source data.}
\label{tab:worked-forecast-output}
\setlength{\tabcolsep}{3pt}
\renewcommand{\arraystretch}{1.18}
\input{tables/same_patient_clinical_io}
\end{table}

\paragraph{What the mask and output probabilities mean.}
Observed FiO$_2$ equals 0.40 and maps to a normalized value of zero with mask
one. Missing MAP also has a zero placeholder, but mask zero distinguishes
it from an observed value. For example, the H1 MAP decoder mean is
79.72 mmHg with predicted observation probability 0.275; no MAP reference
is observed at any displayed horizon. This probability describes availability,
not the probability that 79.72 mmHg is accurate. The decoded mean is an
availability-conditioned output.

\begin{table}[H]
\centering\small
\caption{\textbf{CXR representation outputs from the same rollout.}
RMSE is over 64 train-standardized PCA coordinates, not image pixels.
Persistence repeats the source representation. These are the same image-state
predictions already used in the complete-CXR case; no new image selection is performed.}
\label{tab:same-patient-cxr-output}
\setlength{\tabcolsep}{12pt}
\input{tables/same_patient_cxr_io}
\end{table}

\paragraph{A separate gated-decision output.}
The MIMIC validation action-ranking case in
\cref{fig:mimic-action-ranking-case} is a different patient example.
Among 16 fluid--vasopressor templates, F1--V0 has the lowest imagined mortality
score (13.222\%), versus 13.283\% for F0--V0. Identification evidence is absent,
so the reported branch retains F0--V0. Authorized-set selection is evaluated
in MASS-Syn (\cref{tab:cgrpi}).

%% file: tables/same_patient_clinical_io.tex
\begin{tabular}{@{}lrrrrr@{}}
\toprule
Variable & Source & H1: P / O & H2: P / O & H3: P / O & H4: P / O\\
\midrule
Respiratory rate (/min) & 24.09 & 23.68 / 20.30 & 23.36 / 21.38 & 23.15 / 23.12 & 23.06 / 23.20\\
SpO$_2$ (\%) & 99.33 & 99.04 / 99.83 & 98.74 / 99.33 & 98.48 / 98.67 & 98.28 / 95.67\\
FiO$_2$ (fraction) & 0.40 & 0.40 / 0.40 & 0.40 / 0.40 & 0.40 / 0.40 & 0.40 / 0.40\\
PaO$_2$ (mmHg) & 113.00 & 100.99 / -- & 96.96 / -- & 97.35 / 161.00 & 97.74 / --\\
Creatinine & -- & 1.29 / -- & 1.32 / -- & 1.35 / 0.60 & 1.38 / --\\
Urine output (ml) & 31.67 & 45.09 / 30.83 & 57.67 / 48.75 & 68.89 / 143.33 & 78.46 / 125.00\\
Heart rate (/min) & 107.67 & 105.70 / 82.83 & 103.94 / 92.83 & 102.52 / 101.50 & 101.48 / 116.00\\
MAP (mmHg) & -- & 79.72 / -- & 80.03 / -- & 80.16 / -- & 80.15 / --\\
Lactate & -- & 1.57 / -- & 1.60 / -- & 1.63 / -- & 1.65 / --\\
Hematocrit & -- & 28.10 / -- & 27.52 / -- & 27.04 / 28.60 & 26.65 / --\\
Platelets & -- & 217.71 / -- & 220.57 / -- & 224.20 / 212.00 & 228.82 / --\\
WBC & -- & 12.49 / -- & 12.54 / -- & 12.58 / 9.90 & 12.60 / --\\
Temperature ($^\circ$C) & 37.42 & 37.37 / 37.28 & 37.32 / 37.81 & 37.28 / 37.44 & 37.26 / 37.19\\
\bottomrule
\end{tabular}

%% file: tables/same_patient_cxr_io.tex
\begin{tabular}{@{}lrrrr@{}}
\toprule
CXR representation error & H1 & H2 & H3 & H4\\
\midrule
PulmoWorld & 0.902 & 0.780 & 0.939 & 0.756\\
Persistence & 1.228 & 1.258 & 1.318 & 1.129\\
\bottomrule
\end{tabular}

%% file: appendix/d_integrated_trajectories.tex
\subsection{Clinical case selection and image-state comparison}
\label{app:integrated-trajectories}

\paragraph{Selection of the main-text clinical examples.}
The respiratory-rate, SpO$_2$, and MAP panels in
\cref{fig:asynchronous-evidence} show separate MIMIC patients.
Eligibility requires observed source and H1--H4 values, nonconstant
ground truth, lower four-step MAE for \method{} than each of GRU,
causal Transformer, and persistence, source-inclusive correlation at least
0.8, and MAE divided by the observed range at most 0.25.
Among eligible cases, selection minimizes the worst displayed method's
normalized MAE, rather than maximizing the winning margin.
Nine respiratory-rate, nine SpO$_2$, and one MAP trajectory meet these
outcome-dependent criteria. Their \method{} MAEs are 1.794 breaths/min,
0.630 percentage points, and 1.441 mmHg, respectively.
\Cref{tab:t2-main} reports the locked-test comparison.

\paragraph{Additional trajectory examples.}
Among 13 complete
MASS PaO$_2$ trajectories, one has lower four-step MAE for \method{} than
both the causal Transformer and CLARITY. Its observed H1--H4 values are
68.50, 79.00, 106.00, and 75.00 mmHg, whereas the predictions are
88.74, 89.55, 86.49, and 84.80 mmHg (MAE 15.03 mmHg).
The unobserved source value prevents a persistence comparison.
For MIMIC, median-error examples among \method{} wins over GRU,
causal Transformer, and persistence yield MAEs of 0.98 percentage points
for SpO$_2$ and 4.91 mmHg for MAP. There are 62 such wins among 296
eligible SpO$_2$ trajectories and 18 among 93 eligible MAP trajectories.

\paragraph{A complete CXR sequence.}
The full observed sequence and its same-patient clinical forecasts appear in
\cref{fig:same-patient-io}, with complete numerical outputs in
\cref{tab:worked-forecast-output,tab:same-patient-cxr-output}.
This validation window was selected for complete image coverage rather than
forecast accuracy. Event-window labels are not exact acquisition times;
multiple views contribute to some pooled window representations.
All four methods in the main-text image-error panel are scored in the
same train-standardized, 64-dimensional coordinate system.
CLARITY predictions are transformed from raw PCA coordinates before
scoring; the train mean is zero in this space.
\method{} improves on CLARITY and source persistence in this example.
The comparison concerns representation prediction, not image synthesis.
\FloatBarrier

%% file: appendix/d_case_evolution.tex
\subsection{Imaging and metabolomic evolution}
\label{app:frozen-representation}
\label{app:cases}

Locked-test recursive forecasting scores clinical variables only. Imaging evidence is therefore
reported here through distinct diagnostics: MASS CT/BALF latent errors
and clinical forecast errors using frozen MIMIC CXR encoders
(\cref{tab:imaging}). The subsequent CXR figure separately scores image latents.
Direct VLMs are omitted from the MASS latent panel because they never emit
CT64 or BALF64. The reported scores are representation forecasts for those
modalities.

Together, the following analyses distinguish forecasting a compact modality
representation from recovering an image or identifying an individual
metabolite. Aggregate errors are followed by spatial CT diagnostics, CXR
coverage and error summaries, and longitudinal BALF examples. The latter
also show how sparse repeat sampling limits later-horizon evaluation.

\input{tables/tab_imaging}

\paragraph{Spatial CT diagnostics.}
The CT examples below complement vector-level forecast errors by comparing
observed and predicted token changes. They include a favorable, an intermediate,
and an unfavorable case; representative slices are not anatomically registered.
These displays assess representation dynamics rather than image synthesis.

\begin{figure*}[!htbp]
  \centering
  \includegraphics[width=\textwidth]{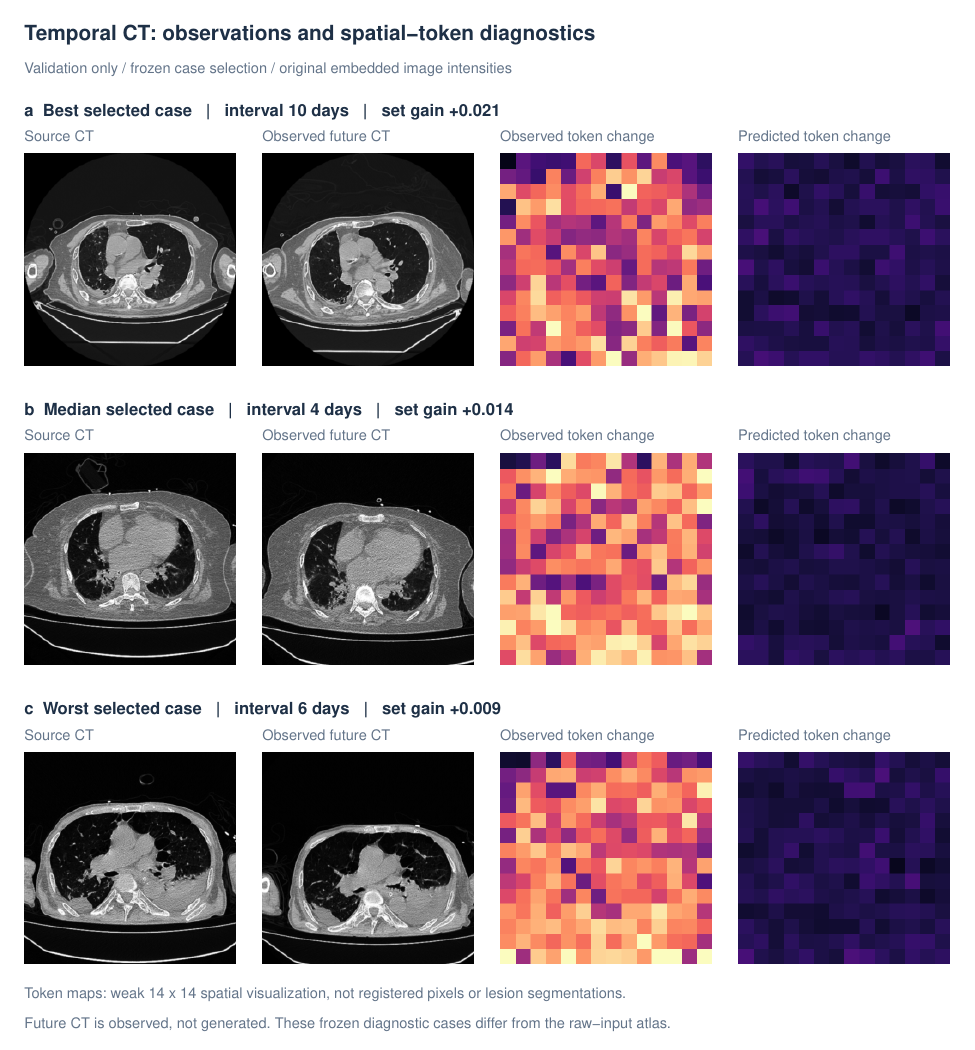}
  \caption{\textbf{MASS CT observations and token-space evolution.}
  Validation CT pairs (best / median / worst set-distance gain after an
  SSIM screen). Columns are current 2D CT, observed future 2D CT, observed
  token change, and predicted token change. Future CT is observed, not generated. No locked-test patient is
  shown. These are the frozen spatial-token diagnostic cases, separate from
  the raw-input atlas.}
  \label{fig:imaging-cases}
  \label{fig:ct-spatial-cases}
\end{figure*}

\begin{figure*}[!htbp]
  \centering
  \includegraphics[width=\textwidth]{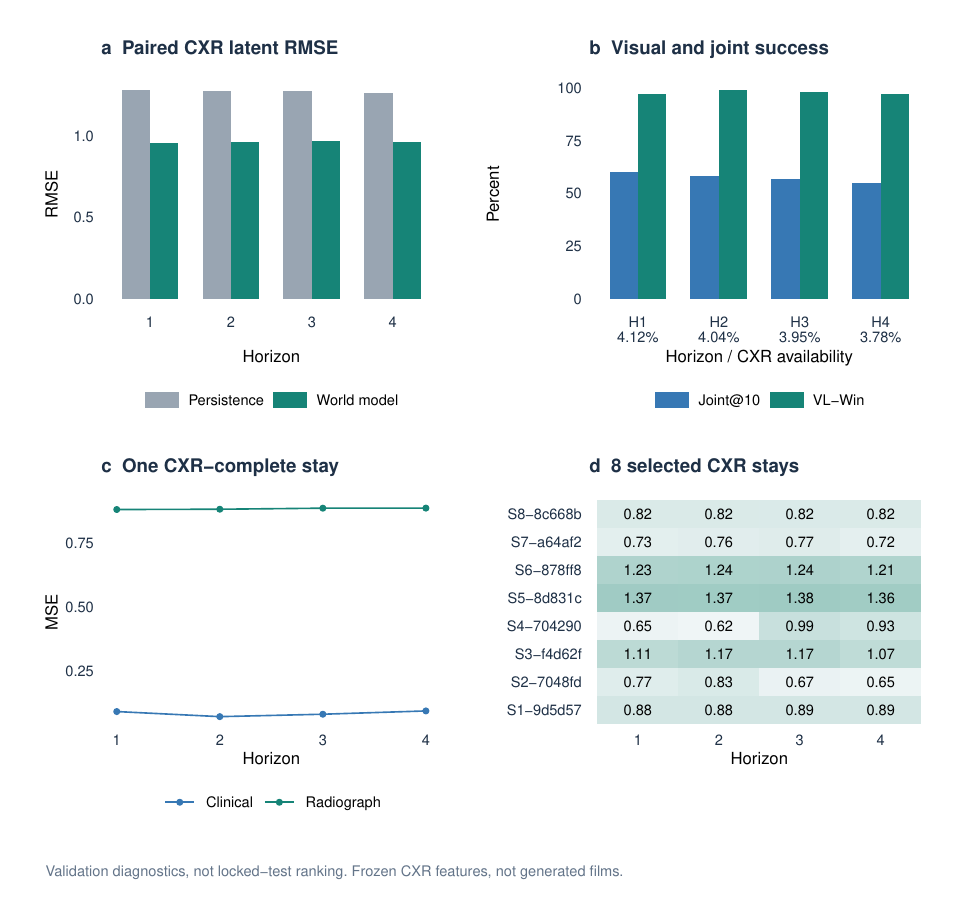}
  \caption{\textbf{MIMIC paired-CXR latent evolution.} Validation
  data, separate from the 300-patient locked test; observed rather than generated
  films. (a)~Frozen Qwen2.5-VL-7B encoder $+$ world model versus source-film
  persistence. (b)~VL-Win is the fraction of source/target-CXR pairs whose
  predicted representation is closer to the target than persistence.
  Joint@10 is the fraction of eligible clinical cells with at most 10\%
  source-normalized error and a simultaneous visual win. The error denominator
  is the larger of the absolute source value and 0.1 times the training IQR
  (with a numerical floor of $10^{-8}$); these rates have
  different denominators. Tick labels give the share of
  windows with a target CXR ($\approx$4\%). (c)~One hashed stay with a
  radiograph at every horizon. (d)~Eight selected low-clinical-MSE stays with CXR at
  H1--H4. Stay labels are pseudonyms, not MIMIC identifiers.}
  \label{fig:mimic-cxr-horizon}
\end{figure*}

\begin{figure}[!htbp]
  \centering
  \includegraphics[width=\textwidth]{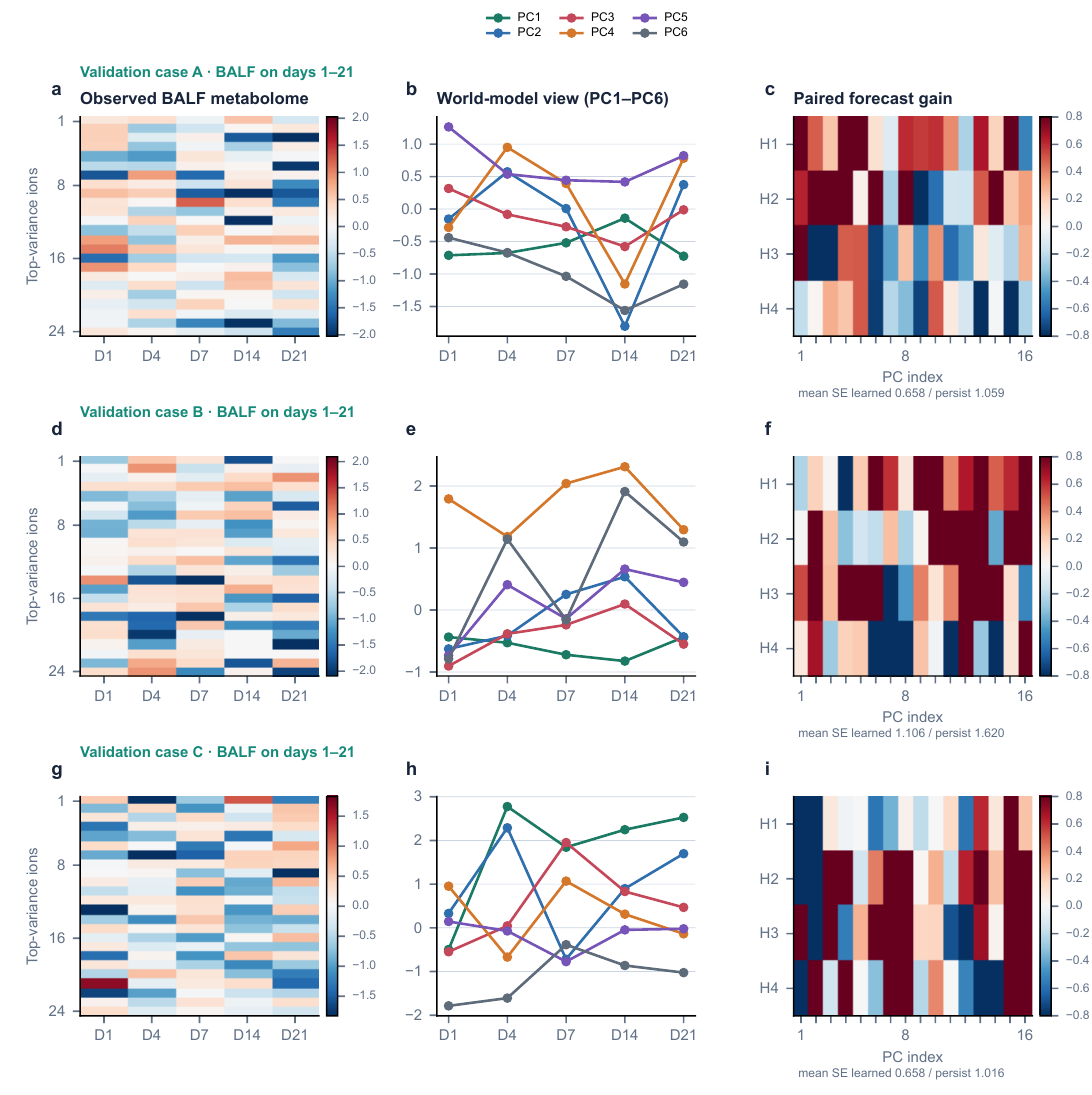}
  \caption{\textbf{Observed BALF metabolomic sequences and predicted state changes.}
  Three development-validation patients with five BALF visits on protocol days
  1/4/7/14/21 (3 of 63 validation patients).
  (a,d,g)~24 highest-variance ions after the train-only log1p+IQR map;
  rows index selected ions and colorbars report transformed intensities.
  (b,e,h)~PCA64 scores for the same visits (PC1--PC6), with a shared legend.
  (c,f,i)~recursive forecasting paired MSE gain versus carry-forward on PC1--PC16; positive means
  \method{} is closer than persistence. The annotations report mean squared
  errors for the learned forecast and persistence. Direct VLMs do not emit BALF64.
  Validation examples only, not the locked-test recursive forecasting ranking.}
  \label{fig:mass-balf-cases}
\end{figure}
\FloatBarrier

%% file: tables/tab_imaging.tex
\begin{table*}[!htbp]
\centering
\caption{\textbf{Development and validation of imaging and metabolomic representations.}
MASS CT64/BALF64 are Clinical19 development rollouts (three-seed mean
standardized MSE; lower is better). Direct VLMs never emit those latents, so
they are omitted rather than shown as dashes. MIMIC reports clinical forecast RMSE using frozen
CXR PCA64 encoders and independently fitted world models on validation
without accessing test patients. These are fitted-encoder comparisons,
not prompted numerical forecasts or the 300-patient locked-test comparison.
These latent forecasts are separate from locked-test recursive forecasting, which uses
the Clinical19 and 13-variable clinical targets. Bold and underline mark the
best and second-best displayed values in each column.}
\label{tab:imaging}
\scriptsize
\setlength{\tabcolsep}{3.4pt}
\renewcommand{\arraystretch}{1.05}

{\footnotesize\textbf{MASS CT64 / BALF64 (development)}}\\[2pt]
\resizebox{0.98\textwidth}{!}{%
\begin{tabular}{@{}lcccccccc@{}}
\toprule
\rowcolor{tblhead}
Method & CT H1$\downarrow$ & CT H2$\downarrow$ & CT H3$\downarrow$ & CT H4$\downarrow$
& BALF H1$\downarrow$ & BALF H2$\downarrow$ & BALF H3$\downarrow$ & BALF H4$\downarrow$ \\
\midrule
GRU & \underline{0.868} & 0.848 & \underline{0.884} & \underline{0.894} & \underline{0.889} & \underline{0.926} & \underline{0.792} & \underline{0.913} \\
Causal Transformer & 0.884 & 0.862 & 0.898 & 0.900 & 0.893 & 0.934 & 0.810 & 0.928 \\
Deterministic WM & 0.895 & 0.879 & 0.941 & 0.986 & 0.897 & 0.964 & 0.847 & 1.000 \\
RSSM & 0.881 & \underline{0.846} & 0.898 & \underline{0.894} & 0.896 & 0.932 & 0.803 & 0.922 \\
\rowcolor{tblours}
\method{} & \textbf{0.836} & \textbf{0.834} & \textbf{0.875} & \textbf{0.881}
& \textbf{0.836} & \textbf{0.921} & \textbf{0.785} & \textbf{0.912} \\
\bottomrule
\end{tabular}%
}

\vspace{5pt}
{\footnotesize\textbf{MIMIC clinical forecasting with frozen CXR encoders (validation RMSE)}}\\[2pt]
\begin{tabular}{@{}llcc@{}}
\toprule
\rowcolor{tblhead}
Model & Source input & Full validation$\downarrow$ & CXR-paired$\downarrow$ \\
\midrule
Persistence & Clinical & 0.8625 & 0.8755 \\
GRU & Clinical & 0.7711 & 0.7456 \\
Causal Transformer & Clinical & 0.7686 & 0.7480 \\
GRU-D & Clinical & 0.9289 & 0.9359 \\
Qwen2.5-VL-7B + fitted WM & Clinical+CXR64 & 0.7653 & 0.7448 \\
Qwen2.5-VL-72B + fitted WM & Clinical+CXR64 & \textbf{0.7651} & \textbf{0.7438} \\
Qwen3-VL-30B-A3B + fitted WM & Clinical+CXR64 & \underline{0.7652} & \underline{0.7442} \\
\bottomrule
\end{tabular}

\vspace{2pt}
\parbox{0.98\textwidth}{\footnotesize
MASS numbers are the CT64/BALF64 block of the Clinical19 development three-seed comparison. MIMIC encoder rows use frozen CXR encoders and a fitted clinical head on validation. They are not the locked 13-variable forecast and they are not Direct-VLM forecasts. Remaining encoder variants sit in
$0.7649$--$0.7654$ on the full validation split. }
\end{table*}

%% file: appendix/d_reproducibility.tex
\section{Reproducibility and Data Access}
\label{app:reproducibility}

\paragraph{Materials and access.}
We will release evaluation code, preprocessing specifications, synthetic
mechanisms, and aggregate results. MIMIC-IV and MIMIC-CXR must be obtained
under the PhysioNet data-use agreement
\citep{johnson2023mimic,johnson2019mimiccxr}; we do not redistribute these
records. MASS was used with permission after personal identifying
information was removed. Its patient-level clinical, CT, and BALF records
are not part of the public release. Any qualified request for controlled
access remains subject to institutional permission and the applicable
data-use conditions.

\paragraph{Evaluation separation.}
The locked MASS and MIMIC tests were evaluated after the respective
protocols were fixed. Test labels were excluded from training, model
selection, calibration, and threshold fitting. Development ablations and
history-enabled prognosis use separate validation populations, as specified
in \cref{app:mechanism-evidence}. Uncertainty estimates resample patients
for clinical comparisons; MASS-Syn policy comparisons instead summarize
independent simulated environments. These sources of uncertainty should
not be interpreted interchangeably.

\paragraph{Scope of reproducibility.}
The appendix specifies cohort construction, model dimensions, loss functions,
training schedules, comparator interfaces, and case-selection criteria.
Clinical forecasting, representation prediction, and synthetic policy
selection remain distinct evaluations: Cap-MBRL is tested on known-mechanism
MASS-Syn, while real-cohort action analyses assess care-conditioned forecasts
and authorization.